\documentclass[letterpaper]{article} 
\usepackage{aaai25}  
\usepackage{times}  
\usepackage{helvet}  
\usepackage{courier}  
\usepackage[hyphens]{url}  
\usepackage{graphicx} 
\usepackage{natbib}  
\usepackage{caption} 
\usepackage{algorithm}
\usepackage{algorithmic}

\usepackage{amsmath}
\usepackage{url}            
\usepackage{booktabs}       
\usepackage{amsfonts}       
\usepackage{nicefrac}       
\usepackage{microtype}      
\usepackage{xcolor}         
\usepackage{graphicx}
\usepackage{booktabs}
\usepackage{caption}
\usepackage{subcaption}
\usepackage{amssymb}
\usepackage{bbding}
\usepackage{pifont}
\usepackage{wasysym}
\usepackage{utfsym}
\usepackage{fontawesome}
\usepackage{makecell}
\usepackage{multirow}                 
\usepackage{float}                    
\usepackage{bigstrut,rotating,booktabs}
\usepackage{array}
\usepackage{arydshln}
\usepackage[capitalize]{cleveref}
\usepackage{bigstrut}
\UseRawInputEncoding

\usepackage{color}

\usepackage{newfloat}
\usepackage{listings}
\DeclareCaptionStyle{ruled}{labelfont=normalfont,labelsep=colon,strut=off} 
\floatstyle{ruled}
\newfloat{listing}{tb}{lst}{}
\floatname{listing}{Listing}
\title{AFFAKT: A Hierarchical Optimal Transport based Method for Affective Facial Knowledge Transfer in Video Deception Detection}
\author{
    Zihan Ji\textsuperscript{\rm 1},
    Xuetao Tian\textsuperscript{\rm 2},
    Ye Liu\textsuperscript{\rm 1}\thanks{The corresponding author.}
}
\affiliations{
    \textsuperscript{\rm 1}School of Future Technology, South China University of Technology, Guangzhou, China\\
    \textsuperscript{\rm 2}Faculty of Psychology, Beijing Normal University, Beijing, China\\
    ftjizihan@mail.scut.edu.cn, xttian@bnu.edu.cn, yliu03@scut.edu.cn

}

\begin{document}

\maketitle

\begin{abstract}
    The scarcity of high-quality large-scale labeled datasets poses a huge challenge for employing deep learning models in video deception detection. To address this issue, inspired by the psychological theory on the relation between deception and expressions, we propose a novel method called AFFAKT in this paper, which enhances the classification performance by transferring useful and correlated knowledge from a large facial expression dataset. Two key challenges in knowledge transfer arise: 1) \textit{how much} knowledge of facial expression data should be transferred and 2) \textit{how to} effectively leverage transferred knowledge for the deception classification model during inference. Specifically, the optimal relation mapping between facial expression classes and deception samples is firstly quantified using proposed H-OTKT module and then transfers knowledge from the facial expression dataset to deception samples. Moreover, a correlation prototype within another proposed module SRKB is well designed to retain the invariant correlations between facial expression classes and deception classes through momentum updating. During inference, the transferred knowledge is fine-tuned with the correlation prototype using a sample-specific re-weighting strategy. Experimental results on two deception detection datasets demonstrate the superior performance of our proposed method. The interpretability study reveals high associations between deception and negative affections, which coincides with the theory in psychology.
\end{abstract}

\section{Introduction}
\label{sec:introduction}
\begin{figure}[tb]    
    \centering
    \includegraphics[width=0.8\linewidth]{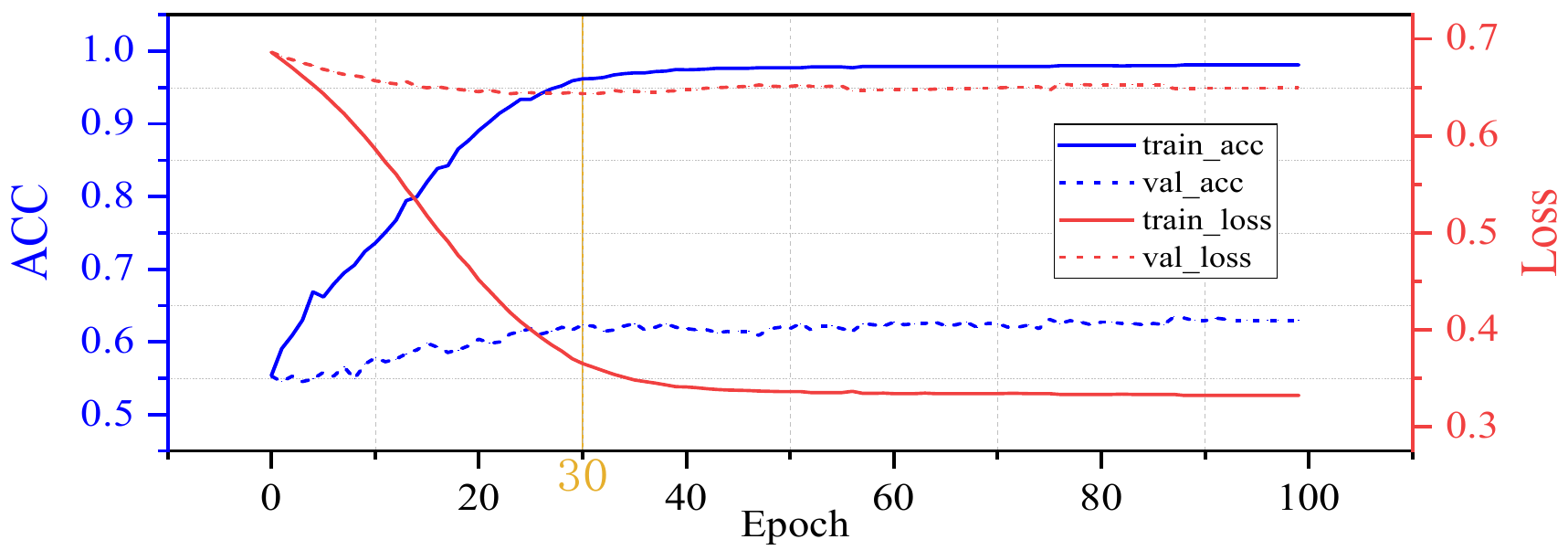}
    \caption{ACC and loss remain unchanged after 30 epochs.
    }
    \label{fig:issue}
    \vspace{-0.4cm}
\end{figure}

Video deception detection has attracted a huge interest in various fields including law enforcement, jurisprudence, national security, business and interviewing \cite{chebbi2023deception}. In the earlier days, researchers proposed several statistic-based methods \cite{jaiswal2016truth, rill2019high, mathur2020introducing}, which utilize facial features, such as OpenFace \cite{baltrusaitis2018openface}, action units \cite{jaiswal2016truth} and face landmarks \cite{rill2019high}, to achieve classification by applying machine learning methods on facial features with statistically significant differences. The performance of these methods heavily relies on the expert knowledge to construct and select valid feature sets. Hereafter, several deep learning based methods \cite{zhang2022fine, hsiao2022attention, guo2023audio} employed different advanced neural networks to automatically extract powerful feature representations for video deception detection. However, their performances highly depends on the availability of high-quality labeled real data \cite{chen2019closer}. Moreover, current datasets, \textit{e.g.}, Real Life Trial (RTL) \cite{perez2015deception}, DOLOS \cite{guo2023audio}, typically contain small number of annotated samples, which poses limitations on training deep neural networks, thereby hindering further performance improvement. Taking PECL(visual) on DOLOS \cite{guo2023audio} as an example (\cref{fig:issue}, refer to appendix for detailed settings), PECL(visual) cannot obtain enough knowledge about detecting deception due to limited data and poor feature representation. Consequently, a key question in deception detection is how to develop a superior deep learning based method when limited labeled deception data is available.

To address this issue, motivated by the insights from psychological theory and previous researches that certain facial movements and expressions are associated with deception \cite{Le_2016, ZUCKERMAN19811, depaulo2003cues, zloteanu2020reconsidering}, we propose a hierarchical optimal transport \cite{guo2022adaptive} based method for AFfective FAcial Knowledge Transfer (AFFAKT). The objective of AFFAKT is to transfer and leverage the knowledge properly from a large-scale video facial expression recognition (VFER) dataset (\textit{source domain}) to enhance video deception detection (\textit{target domain}) model, enabling better detection performance. Subsequently, there are two key challenges when transferring knowledge: 1) \textit{how much} knowledge of facial expression data should be transferred and 2) \textit{how to} effectively leverage transferred knowledge for the deception classification model during inference. 

Two modules are integrated in AFFAKT to handle these two challenges. Firstly, Hierarchical Optimal Transport Knowledge Transfer (H-OTKT) module is devised to automatically quantify the potential correlation between the class of facial expressions and deception samples via hierarchical optimal transport (H-OT). Such correlation mapping would be employed to determine how much knowledge from different categories is transferred to each sample. The transferred knowledge is further integrated with extracted target deception feature to enhance its representation. Secondly, Sample-specific Re-weighting Knowledge Bank (SRKB) module is proposed to learn the invariant class-level relation between VFER and video deception detection datasets via momentum updated correlation prototype whilst training. During testing phase, for each test sample, its correlation mapping with facial expression classes calculated by H-OTKT is fine-tuned using the correlation prototype via a sample-specific re-weighting strategy, so that inaccurate correlation can be alleviated. This strategy facilitates the robustness of the estimated relation mapping, and takes advantage of the learned correlated knowledge for detection inference more efficiently. The experimental results on two video deception detection datasets demonstrate the superior performance of our proposed method. And the interpretability studies reveal high associations between deception and negative affections, which coincide with the theory in psychology. The code can be found at \texttt{https://github.com/Zander-J/AFFAKT}.

The main contributions of this paper can be summarized as follows:
\begin{itemize}
    \item We propose a new method called AFFAKT, which facilitates the classification performance in video deception detection by transferring and leveraging the correlated knowledge from a large-scale facial expression dataset.
    \item An H-OTKT module is presented to quantify the correlation mapping between facial expression classes and deception samples, so as to transfer appropriate amount of knowledge from different source classes to deception samples. Besides, an SRKB module is exploited to enhance the correlation mapping with correlation prototype through a sample-specific re-weighting strategy during inference. SRKB effectively leverages transferred knowledge, thereby improving deception detection accuracy.
    \item Experiments including comparison experiments, ablation studies and interpretability  analysis are performed on two video deception detection datasets. The superior performance validate the effectiveness of our proposed strategies. The interpretability studies reveal high associations between deception behavior and negative affections. 
\end{itemize}

\section{Related Work}
\label{sec:related}
\noindent\textbf{Video Deception Detection. }Video deception detection is one of the main tasks in affective computing and psychological researches \cite{Krishnamurthy2018ADL, perez2015deception, jimaging4100120}. Earlier methods relied on manually selected statistical features including OpenFace \cite{jaiswal2016truth, rill2019high}, action units \cite{jaiswal2016truth, avola2019automatic}, or gestures \cite{csen2020multimodal} to achieve deception detection. Recently, the powerful representation capability of end-to-end deep learning has greatly improved the accuracy of deception detection. \cite{9446553, Ding_2019_CVPR, Krishnamurthy2018ADL,guo2023audio} used ResNet to encode video frames and LSTM to capture temporal information. \cite{guo2023audio} applied transformer \cite{vaswani2017attention} based ViT \cite{dosovitskiy2020image} with adapter tuning \cite{houlsby2019parameter} as temporal encoder to perform deception detection, leading to significant improvement of classification performance. However, one of the key problems is that the high-quality labeled dataset is always limited due to the initiative of lying and the expensive cost of annotation \cite{SANCHEZJUNQUERA2020122}, which seriously hinders the development of deep learning methods. 

\noindent\textbf{Optimal Transport. }Optimal transport (OT) \cite{COTFNT} is a mathematical framework, seeking the most efficient way of transporting one distribution of mass into another. Let \({p}=\sum_{i=1}^n a_i\delta_{\mathbf{X}_{A_i}}\) and \({q}=\sum_{j=1}^m b_j\delta_{\mathbf{X}_{B_j}}\) be \(n\) and \(m\) dimensional discrete probability distributions for two finite sets \(\mathbf{X}_A=\{\mathbf{X}_{A_i}\}_{i=1}^n, \mathbf{X}_B=\{\mathbf{X}_{B_j}\}_{j=1}^m\) respectively, where \(\boldsymbol{a} \in \Delta_n\) and \(\boldsymbol{b} \in \Delta_m\), \(\Delta_n\) and \(\Delta_m\) are the probability simplex of $\mathbb{R}^n$ and $\mathbb{R}^m$, and $\delta_{\mathbf{X}_{*}}$ refers to a point mass located at coordinate $\mathbf{X}_{*}\in\mathbb{R}^{d}$. Denoting \(\mathbf{M}\in\mathbb{R}^{n\times m}_{+}\) as the cost matrix with \(\mathbf{M}_{i,j}=\mathcal{M}(\mathbf{X}_{A_i},\mathbf{X}_{B_j})\), which means the cost to transport one unit of mass between elements of the sets. Then, the transport plan matrix \(\mathbf{T}\) is obtained by solving:
\begin{small}
\begin{align}
\label{eq:stdot}
    \text{OT}({p}, {q})=\min_{\mathbf{T}\in\Pi({p}, {q})}<\mathbf{T}, \mathbf{M}>_{\text{F}}
\end{align}
\end{small}

\noindent where \(<\cdot,\cdot>_{\text{F}}\) is the Frobenius dot-product. The constrain \( \Pi(p,q):=\{\mathbf{T}\in\mathbb{R}^{n\times m}_{+}|\sum_{i}^n\mathbf{T}_{i,j}=b_j, \sum_{j}^m\mathbf{T}_{i,j}=a_i\}\) enforces  \(\mathbf{T}\) to have $p, q$ as its marginals. It should be noted that \(\mathbf{T}\) can be interpreted as the probabilistic correspondence between the elements of $p$ and $q$. If the transport cost $\mathbf{M}_{i,j}$ between $\mathbf{X}_{{A}_i}$ and $\mathbf{X}_{{B}_j}$ is high, then a low correlation \(\mathbf{T}_{ij}\) should be obtained. \cref{eq:stdot} is a linear assignment problem, which is expensive to solve.
Fortunately, a entropy regularized OT  has been developed as follows:
\begin{small}
\begin{align}
    \label{eq:stdotreg}
    \text{OT}({p}, {q})=\min_{\mathbf{T}\in\Pi({p}, {q})}<\mathbf{T}, \mathbf{M}>_{\text{F}}-\epsilon\mathcal{H}(\mathbf{T})
\end{align}
\end{small}

\noindent Here, \(\mathcal{H}(\mathbf{T})=-\mathbf{T}\log\mathbf{T}\) is the entropic regularization. \cref{eq:stdotreg} can be solved by the Sinkhorn algorithm efficiently \cite{cuturi2013sinkhorn}. 

\noindent\textbf{Hierarchical Optimal Transport. }
Hierarchical optimal transport usually contains high-level and low-level OT, where high-level OT learn the optimal transport plan with a given cost matrix, and the given cost matrix depends on the solution of low-level OT. Recently, H-OT has been recently studied for various tasks including multimodal distribution alignment \cite{lee2019hierarchical}, few-shot learning \cite{guo2022adaptive}. For example, in \cite{lee2019hierarchical}, H-OT was used to leverage cluster structure in data to improve alignment in noisy, ambiguous, or multimodal settings. \cite{guo2022adaptive} proposed a novel distribution calibration method for few-show learning, where an adaptive weight matrix representing the relations between the base classes and novel samples is computed by hierarchical optimal transport.

\section{The Proposed Method}
\begin{figure*}[!htbp]  
    \centering
    \includegraphics[width=0.9\linewidth]{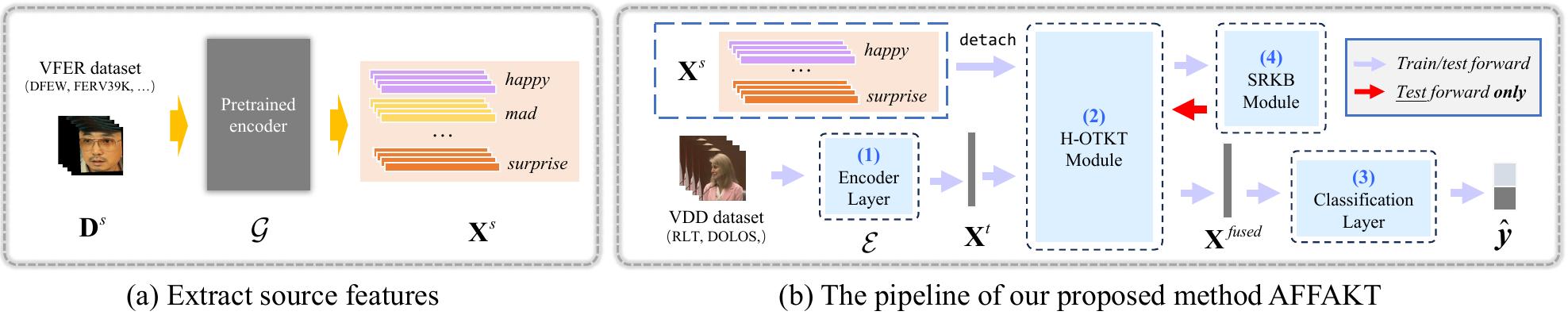}
    \caption{(a) 
    Source features are extracted by a pre-trained encoder in advance. (b) The pipeline of our proposed AFFAKT. Four modules in AFFAKT are in blue background.
    }
    \label{fig:overviewmodel}
\end{figure*}

Our proposed method AFFAKT for video deception detection contains four modules shown in \cref{fig:overviewmodel} (b): (1) Encoder layer, (2) Hierarchical Optimal Transport Knowledge Tranfer (H-OTKT) module, (3) Classification layer, and (4) Sample-specific Re-weighting Knowledge Bank (SRKB) module. Assuming a video deception detection dataset is denoted as \(\mathbf{D}^{t}=\{(\mathbf{V}^t_i, y^t_i)|\mathbf{V}^t_i\in\mathbb{R}^{F\times 3\times H \times W}, y_i^t\in\{1,\dots,L^t\}\}_{i=1}^{N^t}\), where \((\mathbf{V}^t_i, y^t_i)\) is the \(i\)-th video sample and its ground truth label, \(N^{t}\) represents the number of samples in \(\mathbf{D}^{t}\), \(F\) is the number of video frames, \(H\) and \(W\) are the height and width of video frames, \(L^t\) is the number of target categories (\textit{i.e., } deceptive and truthful). Our idea is to improve the classification performance on \(\mathbf{D}^{t}\) by transferring useful and correlated knowledge from a large-scale VFER dataset \(\mathbf{D}^{s}=\{(\mathbf{V}^s_j, y^s_j)\}_{j=1}^{N^s}\), where \(y^s_j\in\{1,\dots,L^s\}\), \(L^s\) is the number of categories in \(\mathbf{D}^{s}\), and $N^s$ is the number of samples in \(\mathbf{D}^{s}\). Each category stands for one expression. In order to utilize $\mathbf{D}^s$ efficiently, a pre-trained encoder \(\mathcal{G}\) is firstly employed to extract VFER feature representation \(\mathbf{X}^s\in\mathbb{R}^{N^s\times d}\) in advance, \textit{i.e.}, \(\mathbf{X}^s=\mathcal{G}(\mathbf{V}^s)\), where \(d\) is the embedding dimension. By grouping \(\mathbf{X}^s\) with the ground truth labels, \(\mathbf{X}^{s}=\{\mathbf{X}^{s,k}\in\mathbb{R}^{J_k\times d}\}_{k=1}^{L^s}\) with $\sum_{k=1}^{L^s}J_k=N^{s}$, where \(\mathbf{X}^{s,k}\) stands for feature embeddings of $J_k$ samples in the \(k\)-th class. This process is shown in \cref{fig:overviewmodel} (a). Pseudo-code of AFFAKT, all the symbol notations and their descriptions used in this paper are summarized in appendix.
\begin{figure}[!b]
  \centering
    \includegraphics[width=0.75\linewidth]{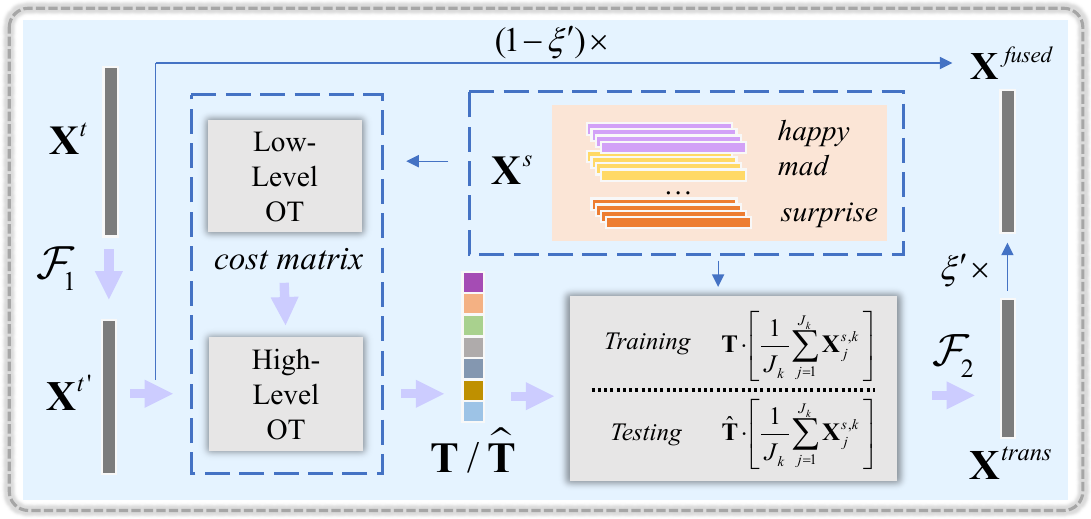}
    \caption{H-OTKT module. It formulates the relation mapping between source classes and target samples, and then performs knowledge transfer.
  }
  \label{fig:HOTKT}
\end{figure}

\subsection{Encoder Layer}
For a video deception detection dataset, widely-used video and audio pre-trained models $\mathcal{E}$, \textit{i.e.}, VideoMAE \cite{tong2022videomae} and W2V2 \cite{baevski2020wav2vec, guo2023audio}, are employed to learn visual and audio features separately. In our settings, there are 12 layers and 4 layers in VideoMAE and W2V2 respectively, where each layer contains a multi-head self-attention module and a feed-forward network. Since the number of samples in deception dataset is too small, it is difficult to fine-tune them directly, even resulting in catastrophic forgetting \cite{houlsby2019parameter}. Inspired by LoRA \cite{hu2021lora}, adapter tuning with UT-Adapter \cite{guo2023audio} is employed. Assume that the feature representations of visual and audio modalities are \(\mathbf{X}^{t}_{(v)}\) and \(\mathbf{X}^{t}_{(a)}\) separately. When multiple modalities including both visual and audio are considered, we simply fuse them with average weights as \(\mathbf{X}^{t}_{(f)}=0.5\mathbf{X}^{t}_{(v)}+0.5\mathbf{X}^{t}_{(a)}\). For simplicity, we use \(\mathbf{X}^{t}\) as the target sample embeddings generated by $\mathcal{E}$ in the following, \textit{i.e.}, $\mathbf{X}^{t}=\mathcal{E}(\mathbf{V}^t)$.

\subsection{Hierarchical Optimal Transport Knowledge Transfer (H-OTKT)}
\label{sec:hotkt}
As presented before, we aim to improve the classification performance by transferring VFER domain knowledge $\mathbf{X}^{s}$ to target deception domain. Since the label space and the distribution of two domains in the feature space are different, one of the key questions is how much knowledge of facial expression data should be transferred. Based on hierarchical optimal transport, we propose H-OT Knowledge Transfer (H-OTKT) module with high-level and low-level OT illustrated in \cref{fig:HOTKT}. 

In particular, high-level OT learn the optimal correlation between classes of VFER dataset and samples of deception dataset with a given cost matrix, where the cost matrix depends on the total low-level OT distance between each target deception sample and all samples from each class of VFER dataset.

\(\mathbf{X}^{t}\) is firstly mapped into \({\mathbf{X}^{t}}'=\mathcal{F}_{1}(\mathbf{X}^{t})\in\mathbb{R}^{n\times d}\) by an MLP \(\mathcal{F}_{1}\), such that the feature spaces between source and target domain could be the same, where \(n\) is the batch size. Let \(\mathcal{Q}=\sum_{k=1}^{L^{s}}\frac{1}{L^{s}}\delta_{\mathcal{Q}^k}\) as the discrete uniform distribution over $L^{s}$ classes of VFER dataset, $\mathcal{Q}^k$ is the representation vector of $k$-th class. And \(\mathcal{P}=\sum_{i=1}^n\frac{1}{n}\delta_{{\mathbf{X}^{t}_i}'}\) is the discrete uniform distribution over \(n\) target deception samples. Then, according to \cref{eq:stdotreg}, the entropic regularized OT between $\mathcal{P}$ and $\mathcal{Q}$ is:
\begin{small}
\begin{equation}
\label{eq:highlevelot}
\begin{aligned}
    \text{OT}_{high}(\mathcal{P}, \mathcal{Q}) = \min_{\mathbf{T}\in\Pi(\mathcal{P}, \mathcal{Q})}<\mathbf{T}, \mathbf{M}>_{\text{F}} - \epsilon\mathcal{H}(\mathbf{T})
\end{aligned}
\end{equation}
\end{small}

\noindent where \(\mathbf{T}\in\mathbb{R}^{n\times L^{s}}\) and \(\mathbf{M}\in\mathbb{R}^{n\times L^{s}}\) are the transport plan and the cost matrix between facial expression classes and target deception samples. Each element \(\mathbf{T}_{i,k}\) indicates the importance of the \(k\)-th class in VFER dataset for the $i$-th sample in deception mini-batch, determining which class and how much of knowledge should be transferred. Besides, \(\mathbf{T}\) should satisfy the following constraint:
\begin{small}
    \begin{equation}
    \begin{aligned}
    \label{eq:highlevelconstraint}
    \Pi(\mathcal{P}, \mathcal{Q}):=\left\{\sum_{i=1}^n\mathbf{T}_{i,k}=\frac{1}{L^{s}},\;\;\sum_{k=1}^{L^{s}}\mathbf{T}_{i,k}=\frac{1}{n}\right\}
\end{aligned}
\end{equation}
\end{small}

It is apparently that the solution $\mathbf{T}$ relies on the cost matrix $\mathbf{M}$, simply applying cosine similarity with the features of samples from deception mini-batch and the mean of features from each class of VFER dataset may lead to sub-optimal solution. Moreover, the contribution of different samples in each class may be various. So, we utilize another optimal transport formulation to obtain the optimal \(\mathbf{M}\). According to \cite{guo2022adaptive}, the empirical distribution of the $k$-th class is expressed as \(\mathcal{Q}^{k}=\sum_{j=1}^{J_k}p_j^k\delta_{\mathbf{X}^{s,k}_j}\), where the importance \(p^k_j\) of the \(j\)-th sample in the \(k\)-th source class is obtained by logistic regression score. Therefore, a low-level entropic regularized OT is further defined as follows: 
\begin{small}
\begin{equation}
\begin{aligned}    
    \label{eq:lowlevelot}
    &\text{OT}_{low}(\mathcal{P}, \mathcal{Q}^{k}) = \\
    &\min_{\mathbf{T}^{low,k}\in\Pi(\mathcal{P}, \mathcal{Q}^{k})}<\mathbf{T}^{low,k}, \mathbf{M}^{low,k}>_{\text{F}} - \epsilon \mathcal{H}(\mathbf{T}^{low,k})
\end{aligned}
\end{equation}
\end{small}

\noindent \(\Pi(\mathcal{P}, \mathcal{Q}^{k}):=\left\{\sum_j^{J_k}\mathbf{T}^{low,k}_{i,j}p^k_j=\frac{1}{n},\sum_i^n\mathbf{T}^{low,k}_{i,j}\frac1n=p^k_j\right\}\) is the constrain, and \(\mathbf{T}^{low,k}\) is the transport plan between each sample in mini-batch and samples in the \(k\)-th source domain class. \(\mathbf{M}^{low,k}\in\mathbb{R}^{n\times J_k}\) is determined by cosine similarity, \textit{i.e.}, \(\mathbf{M}^{low,k}_{i,j}=1-\cos({\mathbf{X}^{t}_i}',\mathbf{X}^{s,k}_{j})\). The cost matrix $\mathbf{M}$ in high-level OT of \cref{eq:highlevelot} will be replaced by the total OT distance between each target deception sample and all sample in each class of VFER dataset, \textit{i.e.}, $\mathbf{M}_{:,k}=<\mathbf{T}^{low,k},\mathbf{M}^{low,k}>_{\text{F}}$. 

For the optimization, both \cref{eq:lowlevelot} and \cref{eq:highlevelot} are solved by Sinkhorn algorithm \cite{cuturi2013sinkhorn} hierarchically. Using the OT distance calculated from low-level OT as the cost \(\mathbf{M}\) of high-level OT adaptively, H-OTKT is able to obtain the transport weight \(\mathbf{T}\) between deception samples and facial expression classes, which is the potential correlation mapping of facial expression classes for target samples.

Once we obtained correlation mapping $\mathbf{T}$ by solving \cref{eq:highlevelot}, knowledge transformation can be performed. For each sample in deception domain, more knowledge from highly associated classes should be transferred, while knowledge from uncorrelated classes should not be transferred. To realize it, the transferred knowledge $\mathbf{X}^{trans}\in\mathbb{R}^{n\times d}$ is represented as follows:
\begin{small}
\begin{equation}
    \begin{aligned}
    \label{eq:transfer}
    \mathbf{X}^{trans}_i=    \mathcal{F}_{2}\left(n\cdot\sum_{k=1}^{L^s}\mathbf{T}_{i,k}\left[\frac{1}{J_k}\sum_{j=1}^{J_k}\mathbf{X}^{s,k}_{j}\right]\right),
    \;\;
    i=1,\cdots,n
\end{aligned}
\end{equation}
\end{small}

\noindent where  $\frac{1}{J_k}\sum_{j=1}^{J_k}\mathbf{X}^{s,k}_{j}$ denotes the average feature of samples belonging to the $k$-th class in source domain; $\mathbf{T}_{i,k}$ quantifies the correlation weight between the $k$-th source class and $i$-th deception sample; \(n\) is used for scaling due to the constraint in high-level OT. And \(\mathcal{F}_{2}\) is an MLP. In order to integrate the transferred knowledge $\mathbf{X}^{trans}$ with features $\mathbf{X}^{t'}$ extracted from target samples, the fused representation of deception detection samples are calculated as: 
\begin{equation}
    \begin{aligned}
\label{eq:fuse}
\mathbf{X}^{fused}=\xi'\mathbf{X}^{trans}+(1-\xi')\mathbf{X}^{t'}
\end{aligned}
\end{equation}
where \(\xi'\) is the weight of transferred feature \(\mathbf{X}^{trans}\). Since it's hard to learn excellent \(\mathbf{X}^{t'}\) at the beginning of the training phase, a curriculum learning strategy \cite{kumar2010self, wang2021survey} is adopted as \(\xi'=\frac{\xi}{2}\times\left(1-\cos\left(\frac{e-1}{N_e}\times\pi\right)\right)\), where \(e\) is the current training epoch number and \(N_e\) is the total training epoch number. As \(\xi'\) is gradually increased, a better \(\mathbf{X}^{t'}\) is gained for H-OTKT.

\subsection{Classification Layer}
\label{sec:clsandloss}
The final classification layer contains one MLP with softmax, which takes \(\mathbf{X}^{fused}\) as input and outputs the predicted label \(\hat{\boldsymbol{y}}\in\mathbb{R}^{n\times L^t}\):
\begin{align}
    \hat{\boldsymbol{y}} = \mathcal{F}_3(\mathbf{X}^{fused})
\end{align}
Here, \(\mathcal{F}_3\) is the MLP classifier. With ground truth label \(\boldsymbol{y}^{t}=[y^t_1, \dots, y^t_n]\), the classification loss function is formulated as:
\begin{align}
    \label{eq:celoss}
    \mathcal{L}_{ce}(\boldsymbol{y}^{t}, \hat{\boldsymbol{y}})=-\mathbb{E}_{\boldsymbol{y}^{t}}[\log{\hat{\boldsymbol{y}}}]
\end{align}

\noindent where \(\mathbb{E}\) is expectation. To reduce the difference between distribution spaces from source and target domain in H-OTKT, and further improve the final prediction, another loss function is defined based on the Sinkhorn divergence \cite{feydy2019interpolating} to obtain the space discrepancy between class average of \(\mathbf{X}^{s}\) and \({\mathbf{X}^{t}}'\) \cite{nguyen2022improving}:
\begin{small}
\begin{equation}    
    \label{eq:otloss}
    \scalebox{0.9}{
    $\mathcal{L}_{ot}(\mathbf{X}^{t'}, \mathbf{X}^{s}) = ds_{\text{OT}}(\mathcal{P}, \mathcal{Q}) - \frac12ds_{\text{OT}}(\mathcal{P}, \mathcal{P}) - \frac12ds_{\text{OT}}(\mathcal{Q}, \mathcal{Q})$
    }
\end{equation}
\end{small}

\noindent where $ds_{\text{OT}}(\cdot,\cdot)$ is the total OT cost between two distributions solved by the regular OT (\cref{eq:stdot}) with cosine similarity as cost function. Then the total loss function is formulated as:
\begin{align}
    \label{eq:loss}
    \mathcal{L} = \mathcal{L}_{ce}+\eta\mathcal{L}_{ot}
\end{align}

In \cref{eq:loss}, the $\mathcal{L}_{ce}$ term optimizes the whole network to improve the classification performance while the $\mathcal{L}_{ot}$ term is used for reducing the discrepance between the source feature space and the target feature space.

\subsection{Sample-specific Re-weighting Knowledge Bank (SRKB) Module}
\label{sectionSRKB}
\begin{figure}[htb]  
    \centering
    \includegraphics[width=0.9\linewidth]{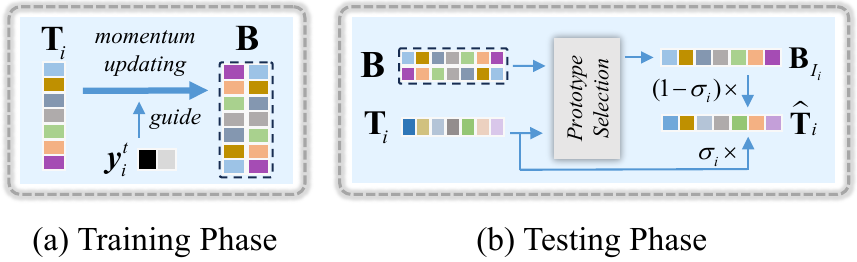}
    \caption{SRKB module. (a) Training phase: \(\mathbf{B}\) is momentum updated to maintain the invariant knowledge of each target class relation with source classes; (b) Testing phase, SRKB module uses the learned \(\mathbf{B}\) and sample-specific re-weighting strategy to enhance the detection performance.
    }
    \label{fig:SRKB}
\end{figure}

Optimal knowledge from proper classes in facial expression dataset has been obtained by H-OTKT. Empirically, samples in deception dataset have varying semantics, thus the class relation would be various because of random data sampling \cite{wang2020learning}. In order to more efficiently use learned knowledge from H-OTKT, and further improve the robustness during testing phase, a plug-in Sample-specific Re-weighting Knowledge Bank (SRKB) module with no additional trainable parameters shown in \cref{fig:SRKB} (a) and (b) is constructed.

Aiming at obtaining more robust and general effective information from VFER dataset and eliminate the randomness caused by data sampling \cite{wang2020learning}, a correlation prototype \(\mathbf{B}=[\mathbf{B}_1^{\top}\dots\mathbf{B}_{L^t}^{\top}]^{\top}\in\mathbb{R}^{L^t\times L^s}\) is constructed to store the robust category relation between two domains. \(\mathbf{B}\) is initialized by \(\frac{1}{L^{s}}\) to ensure \(\sum_k^{L^{s}}\mathbf{B}_{l,k}=1, l=1,\dots,L^{t}\). The \(l\)-th row of \(\mathbf{B}\) demonstrates the association weights between the \(l\)-th class of target domain and each class of source domain. As shown in \cref{fig:SRKB} (a), during training phase, momentum updating \cite{laine2016temporal} is introduced to update the correlation prototype \(\mathbf{B}\) by:
\begin{equation}
\scalebox{0.9}{
$
    \begin{aligned}
    &\mathbf{B}_{l}=\alpha\mathbf{B}_{l}+(1-\alpha)\frac{1}{\sum_i^n\mathbb{I}_{\boldsymbol{y}^{t}_i=l}}\sum_{i=1}^n\mathbf{T}_{i}\mathbb{I}_{\boldsymbol{y}^{t}_i=l}\\&l=1,\cdots,L^{t}
    \end{aligned}    $
    }
    \label{eq:emaupdate}
\end{equation}

\noindent where \(\mathbb{I}_{\boldsymbol{y}^{t}_i=l}=1\) if and only if the label of $i$-th deception sample  $\boldsymbol{y}^{t}_i$ equals to $l$, otherwise \(\mathbb{I}_{\boldsymbol{y}^{t}_i=l}=0\). And \(\alpha\) is the momentum factor. In \cref{eq:emaupdate}, the relation between facial expression classes and each target categories (\textit{i.e.}, truthful and deceptive) is accumulated into the correlation prototype \(\mathbf{B}\) with the guidance of label \(\boldsymbol{y}^{t}\). The underlying invariant relation between all deceptive samples and the source domain expressions is preserved.

According to the previous results of affective computing field \cite{rill2019high} and psychology \cite{depaulo2003cues}, deception should  have high correspondences with some specific facial expression, such as \textit{fear}, and \textit{sad}. Ideally, for the \(i\)-th target deception sample, \(\mathbf{T}_{i}\) should be sparse, where several elements are much higher than the others. This indicates the standard deviation of a sparse \(\mathbf{T}_{i}\) should be larger if \(\mathbf{T}_{i}\) is valid, otherwise it could be smaller. When an unsatisfactory transport plan \(\mathbf{T}_{i}\) with small standard deviation is obtained, it is more reliable to use the corresponding correlation prototype \(\mathbf{B}_{I_i}\)  to quantify the correlation mapping instead, where \(I_i\) is the category that \(\mathbf{T}_i\) may belong to. Otherwise, calculated \(\mathbf{T}_i\) could be improved by its corresponding correlation prototype \(\mathbf{B}_{I_i}\). In our experiment, \(\mathbf{B}_{I_i}\) has smallest distance with \(\mathbf{T}_i\) among \(\{\mathbf{B}_1\dots\mathbf{B}_{L^t}\}\). Mathematically, it can be formulated as:
\begin{equation}
\scalebox{0.9}{
$
    \begin{aligned}    
    &\hat{\mathbf{T}}_{i}=\sigma_i\mathbf{T}_{i}+(1-\sigma_i)\mathbf{B}_{I_i}
    \\s.t.\;\;
    &\sigma_i = \left\{
        \begin{aligned}
            0&, &\text{std}(\mathbf{T}_{i})<\nu\\
            \text{std}(\mathbf{T}_{i})-\nu&,&\text{otherwise}
        \end{aligned}
    \right.
    \end{aligned}
    $}
    \label{eq:transportfuse}
\end{equation}

\noindent where \(\text{std}(\cdot)\) is standard deviation function, and \(\nu\) is a threshold. When the standard deviation of \(\mathbf{T}_{i}\) is greater than \(\nu\), we believe H-OTKT module finds a valid transport plan \(\mathbf{T}\) to measure the importance between categories in source domain and samples in target domain. Otherwise, H-OTKT fails to find a valid transport plan due to noise, and \(\mathbf{T}_{i}\) would not be used for transferring knowledge. Note that the standard deviation would always less than \(1\) according to \cref{eq:highlevelconstraint}. During the testing phase, \(\mathbf{T}\) in \cref{eq:transfer} will be replaced by the obtained \(\hat{\mathbf{T}}\in\mathbb{R}^{n\times L^{s}}\) in \cref{eq:transportfuse}.

\section{Experiments}
\begin{table*}[htbp]
  \centering
  \begin{subtable}{0.49\linewidth} 
    \renewcommand\arraystretch{1.175}
  \centering
  \scalebox{0.57}{
    \begin{tabular}{c|c|ccc|ccc}
    \hline
    \multicolumn{2}{c|}{\textbf{Target}} & \multicolumn{3}{c|}{\textbf{RLT}} & \multicolumn{3}{c}{\textbf{DOLOS}}\\
    \hline
    \textbf{Method} & \textbf{Source} & \textbf{F1 score} & \textbf{ACC} & \textbf{AUC} & \textbf{F1 score} & \textbf{ACC} & \textbf{AUC} \\
    \hline
    OpenFace + SVM  & -     & {0.2253} & {0.5293} & {0.5571} & {0.6975} & {0.5355} & {0.5430}\\
    OpenFace + Decision Tree  & -     & {0.5553} & {0.5303} & {0.5303} & {0.5358} & {0.5058} & {0.5058} \\
    OpenFace + Random Forest  & -     & {0.6033} & {0.6033} & {0.5997} & {0.6175} & {0.5367} & {0.5466} \\
    OpenFace + AdaBoost & -     & {0.5199} & {0.5303} & {0.5766} & {0.5536} & {0.5057} & {0.5035}\\
    \hline
    AU + SVM  & -     & {0.4562} & {0.5043} & {0.4670} & {0.6813} & {0.5276} & {0.5242}\\
    AU + Decision Tree  & -     & {0.4466} & {0.4643} & {0.4643} & {0.5453} & {0.5173} & {0.5173} \\
    AU + Random Forest & -     & {0.5534} & {0.5463} & {0.5330} & {0.5808} & {0.5045} & {0.5157} \\
    AU + AdaBoost  & -     & {0.5130} & {0.4877} & {0.4835} & {0.5295} & {0.4876} & {0.4735} \\
    \hline OpenFace + LSTM & -     & {0.5241} & {0.5623} & {0.5952} & {0.5928} & {0.5628} & {0.5854}\\
    AU + LSTM  & -     & {0.4888} & {0.6197} & {0.6760} & {0.6343} & {0.5646} & {0.5868} \\
    ResNet18 + LSTM & -     & {0.4996} & {0.6117} & {0.6387} & {0.6415} & {0.5972} & {0.5668}\\
    \hline
    PECL(only visual) & -     & {0.5880} & {0.6528} & {0.6734} & {0.7010} & {0.6387} & {0.6770} \\
    \hline
    \multirow{3}[2]{*}{FreeLunch } & DFEW  & {0.7612} & {0.8090} & {0.8712} & {0.6961} & {0.6222} & {0.6444} \\
          & FERV39K & {0.7536} & {0.8173} & {0.8677} & {0.6831} & {0.6228} & {0.6456} \\
          & MAFW  & {0.7663} & {0.8173} & {0.8633} & {0.6695} & {0.6155} & {0.6459} \\
    \hline
    \multirow{3}[2]{*}{ADC } & DFEW  & {0.7793} & {0.8173} & {0.8674} & {0.6880} & {0.6716} & {0.7206} \\
          & FERV39K & {0.7667} & {0.8173} & {0.8677} & {0.6830} & {0.6693} & {0.7156} \\
          & MAFW  & {0.7667} & {0.8173} & {0.8664} & {0.6938} & {0.6684} & {0.7180} \\
    \hline
    \multirow{3}[2]{*}{Cr-KD-NCD } & DFEW  & {0.6957} & {0.7200} & {0.6928} & {0.5850} & {0.6091} & {0.6013} \\
          & FERV39K & {0.7805} & {0.7200} & {0.6464} & {0.6720} & {0.5879} & {0.5363} \\
          & MAFW  & {0.7778} & {0.6800} & {0.6368} & {0.7056} & {0.5697} & {0.5427} \\
    \hline
    \multirow{3}[2]{*}{AFFAKT (ours)} & DFEW  & \textbf{{0.8760}} & \textbf{{0.8670}} & \textbf{{0.8789}} & {0.7054} & \textbf{{0.6764}} & \textbf{{0.7212}} \\
          & FERV39K & {0.8277} & {0.8340} & {0.8415} & \textbf{0.7102} & {0.6746} & {0.7203} \\
          & MAFW  & {0.8524} & {0.8500} & {0.8625} & 0.6948 & {0.6612} & {0.6970}\\
    \hline
    \end{tabular}%
    }
    \subcaption{Results with visual modality. }
    \end{subtable}
    \begin{subtable}{0.49\linewidth} 
    \centering
        \scalebox{0.55}{
        \begin{tabular}{c|c|ccc|ccc}
        \hline
        \multicolumn{2}{c|}{\textbf{Target}} & \multicolumn{3}{c|}{\textbf{RLT}} & \multicolumn{3}{c}{\textbf{DOLOS}} \\
        \hline
        \textbf{Method} & \textbf{Source} & \textbf{F1 score} & \textbf{ACC} & \textbf{AUC} & \textbf{F1 score} & \textbf{ACC} & \textbf{AUC} \\
        \hline
        MFCC + MLP  & -     & 0.5226& 0.6367& 0.7030& 0.5963  & 0.5810  & 0.6134   \\
        OpenSMILE + MLP & -     & 0.6885  & 0.6597  & 0.5926  & 0.6867  & 0.5537  & 0.5325   \\
        W2V2 + MLP  & -     & 0.6117  & 0.6780  & 0.6106  & 0.4383  & 0.5421  & 0.5369   \\
        \hline
        PECL(only audio) & -     & 0.7121  & 0.7100  & 0.6962  & 0.6777  & 0.6119  & 0.6281   \\
        \hline
        \multirow{3}[1]{*}{FreeLunch} & DFEW  & 0.6396  & 0.6767  & 0.6869  & 0.6437  & 0.5864  & 0.6157   \\
          & FERV39K & 0.6432  & 0.6850  & 0.6944  & 0.6589  & 0.5979  & 0.6194   \\
          & MAFW  & 0.6402  & 0.6767  & 0.6885  & 0.6490  & 0.5991  & 0.6196   \\
        \hline
        \multirow{3}[1]{*}{ADC} & DFEW  & 0.6402  & 0.6767  & 0.6858  & 0.6196  & 0.6058  & 0.6040   \\
          & FERV39K & 0.6402  & 0.6767  & 0.6858  & 0.6165  & 0.6052  & 0.6039   \\
          & MAFW  & 0.6272  & 0.6767  & 0.6842  & 0.6129  & 0.6046  & 0.6039   \\
        \hline
        \multirow{3}[0]{*}{AFFAKT (ours)} & DFEW  & 0.7267  & 0.7270  & 0.7218  & 0.6822  & 0.6198  & \textbf{0.6391} \\
              & FERV39K & \textbf{0.7316} & 0.7017  & 0.6917 & \textbf{0.6982} & 0.6173  & 0.6385   \\
              & MAFW  & 0.7266  & \textbf{0.7440} & \textbf{0.7396} & 0.6736 & \textbf{0.6198} & 0.6387   \\
            \hline
        \end{tabular}%
        }
        \subcaption{Results with audio modality. }
        \smallskip
    \centering
    \scalebox{0.52}{
        \begin{tabular}{c|c|ccc|ccc}
        \hline
        \multicolumn{2}{c|}{\textbf{Target}} & \multicolumn{3}{c|}{\textbf{RLT}} & \multicolumn{3}{c}{\textbf{DOLOS}} \\
        \hline
        \textbf{Method} & \textbf{Source} & \textbf{F1 score} & \textbf{ACC} & \textbf{AUC} & \textbf{F1 score} & \textbf{ACC} & \textbf{AUC} \\
        \hline
        OpenFace \(\oplus\) OpenSMILE & -     & 0.6895  & 0.6781  & 0.6212  & 0.6124  & 0.5986  & 0.5863   \\
        ResNet18 \(\oplus\) OpenSMILE  & -     & 0.6283  & 0.6853  & 0.6598  & 0.5863  & 0.6152  & 0.6485   \\
        \hline
        PECL & -     & 0.7102  & 0.6939  & 0.7424  & 0.7084  & 0.6597  & 0.6353   \\
        \hline
        \multirow{3}[1]{*}{FreeLunch} & DFEW  & 0.7473  & 0.8010  & 0.8497  & 0.6686  & 0.6258  & 0.6628   \\
              & FERV39K & 0.7460  & 0.8010  & 0.8504  & 0.6504  & 0.6204  & 0.6574   \\
              & MAFW  & 0.7695  & 0.8093  & 0.8547  & 0.6807  & 0.6289  & 0.6669   \\
        \hline
        \multirow{3}[1]{*}{ADC} & DFEW  & 0.7493  & 0.8093  & 0.8446  & 0.6997  & 0.6746  & 0.7307   \\
              & FERV39K & 0.7493  & 0.8093  & 0.8435  & 0.6819  & 0.6729  & 0.7295   \\
              & MAFW  & 0.7493  & 0.8010  & 0.8411  & 0.6976  & 0.6741  & 0.7274   \\
        \hline
        \multirow{3}[0]{*}{AFFAKT (ours)} & DFEW  & 0.8162  & 0.8180  & 0.8381  & 0.7073  & \textbf{0.6810} & 0.7226   \\
              & FERV39K & 0.7946  & 0.8010  & 0.8357  & \textbf{0.7149} & 0.6774  & \textbf{0.7289} \\
              & MAFW  & \textbf{0.8412} & \textbf{0.8427} & \textbf{0.8563} & 0.7111  & 0.6780  & 0.7181   \\
              \hline
                \end{tabular}%
            }
            \label{tab:comparisonva}%
            \subcaption{Results with fused modalities.
            }
        \end{subtable}
      \caption{Comparison results on RTL \cite{perez2015deception} and DOLOS \cite{guo2023audio} dataset with F1 score, ACC and AUC metrics. DFEW \cite{jiang2020dfew}, FERV39K \cite{wang2022ferv39k} and MAFW \cite{liu2022mafw} are source domains.}  
  \label{tab:comparison}%
\end{table*}%

\subsection{Comparison Methods}
We make comparisons with several deception detection methods on RTL \cite{perez2015deception} and DOLOS \cite{guo2023audio} in visual, audio and fused modalities to validate AFFAKT. Three \textit{in-the-wild} VFER datasets are included, \textit{i.e.}, DFEW \cite{jiang2020dfew}, FERV39K \cite{wang2022ferv39k}, MAFW \cite{liu2022mafw}. Refer to appendix for more details about experimental settings and datasets.

For machine learning based methods, visual (OpenFace and action units (AU)) and acoustic (MFCC and OpenSMILE) features are firstly selected, several widely-used classifier including SVM, Decision Tree, \textit{etc.} are applied on visual features for classification and MLP is applied on acoustic features \cite{mathur2020introducing,  avola2019automatic, yang2021multimodal}. For deep learning based approaches including ResNet18+LSTM \cite{9446553, Ding_2019_CVPR, guo2023audio}, W2V2+MLP \cite{guo2023audio,9446553,Krishnamurthy2018ADL}, ResNet18\(\oplus\)OpenSMILE \cite{Krishnamurthy2018ADL,guo2023audio} and PECL \cite{guo2023audio} are tested.

For transfer learning based methods, FreeLunch \cite{yang2021free} and adaptive distribution calibration (ADC) \cite{guo2022adaptive} are adopted to obtain distance and the optimal transport plan between source class and target samples respectively, measuring the quantity $\mathbf{T}$ of knowledge from source classes to target samples. Then perform knowledge transfer via \cref{eq:transfer} and \cref{eq:fuse}. PECL \cite{guo2023audio} is based on adapter-tuning, which is also a transfer learning based method. Another knowledge distillation based method Cr-KD-NCD \cite{gu2023class} is also conducted, where the VFER classes are the known classes and deception classes are treated as novel classes. In this case, we employ the same visual encoder in AFFAKT as the backbone and only visual modality is evaluated. We refer readers to appendix for more details of these methods.

\subsection{Comparison Results}
\label{sec:comparison}
Comparison results on both datasets with visual, audio and fused modality in terms of F1 score, ACC and AUC metrics are shown in \cref{tab:comparison}. We only report the average value of different folds in \cref{tab:comparison} in the main text and the standard deviation between different folds are reported in the appendix.

Overall, our proposed method obtains the best performance across all evaluation metrics, which suggests that our proposed method is effective and advanced for deception detection. Besides, deep learning based models have achieved better results compared to machine learning based models. Such results indicate that deep learning model exhibits better feature extraction ability than traditional methods. Moreover, compared with \cite{guo2023audio}, we can claim that extra facial expression knowledge is helpful to deception detection. AFFAKT has higher classification accuracy than transfer learning based methods \cite{yang2021free,guo2022adaptive}, demonstrating that better knowledge of facial expression data could be transferred and leveraged in our method. On the other hand, compared to larger DOLOS dataset, AFFAKT outperforms deep learning based methods on smaller RLT more significantly, indicating that AFFAKT is able to show better detection performance on datasets with fewer samples, while other deep learning methods fail when given limited labeled deception data. The results vary across different facial expression datasets, since three facial expression datasets contain different expression categories, and different pre-trained encoders $\mathcal{G}$ are employed for each dataset, resulting in different representation abilities in VFER feature representation $\mathbf{X}^s$.

In \cref{tab:comparison} (b) and (c), we can find that AFFAKT achieves the best performance on both deception dataset in both audio and fused modalities. Specifically, for audio modality, the highest ACCs and F1 scores are achieved when the source domain is MAFW and FERV39K, respectively. The highest AUC is achieved on RLT and DOLOS when the source domains are MAFW and DFEW, respectively. For the fused modality, the best performance is achieved on RLT using MAFW as source domain. And on DOLOS, the best F1 score, ACC and AUC are obtained using FERV39K, DFEW and FERV39K as source domain. The discrepancy of source domain between different deception dataset and modalities would be caused by the robustness and the performance of pre-trained \(\mathcal{G}\) as we have discussed above.
\subsection{Ablation Studies}
Ablation studies are conducted to verify H-OTKT and SRKB modules proposed in this paper. As our method contains four modules shown in Fig. \ref{fig:overviewmodel}: Encoder layer, H-OTKT module, classification layer, and SRKB module, the baseline method denoted as Case \underline{A} only includes encoder layer and classification layer, where the classification layer is applied on extracted features $\mathbf{X}^t$ of deception data by encoder layer $\mathcal{E}$ directly. And the baseline model is only trained with classification loss Eq. \eqref{eq:celoss}. 

\begin{table}[b]
  \setlength{\tabcolsep}{1.5mm}
  \centering
  \scalebox{0.62}{    
    \begin{tabular}{c|ccc|c|cc|cc|cc}
    \hline
    \multirow{2}{*}{\textbf{Case}} & \multicolumn{3}{c|}{\multirow{1}{*}{\textbf{Method}}} & \multirow{2}{*}{\textbf{Source}} & \multicolumn{2}{c|}{\textbf{Visual}} & \multicolumn{2}{c|}{\textbf{Audio}} & \multicolumn{2}{c}{\textbf{Fused}} \\
\cline{6-11}   & \textbf{\ding{172}} & \textbf{\ding{173}} & \textbf{\ding{174}}         &       & \multicolumn{1}{c}{\textbf{RLT}} & \multicolumn{1}{c|}{\textbf{DOLOS}} & \multicolumn{1}{c}{\textbf{RLT}} & \multicolumn{1}{c|}{\textbf{DOLOS}} & \multicolumn{1}{c}{\textbf{RLT}} & \multicolumn{1}{c}{\textbf{DOLOS}} \\
    \hline
    A&\ding{55}     & \ding{55}     & \ding{55}     & -     & 0.7840  & 0.6647  & 0.7100  & 0.6119  & 0.7843  & 0.6794   \\
    \hline
    \multirow{3}[2]{*}{B}&\multirow{3}[2]{*}{\checkmark}  & \multirow{3}[2]{*}{\ding{55}} & \multirow{3}[2]{*}{\ding{55}} & DFEW  & 0.8093  & 0.6677  & 0.6937  & 0.6186  & 0.8013  & 0.6804   \\
          &   &       &       & FERV39K & 0.8177  & 0.6646  & 0.6940  & 0.6143  & 0.7763  & 0.6610   \\
          &    &       &       & MAFW  & 0.8010  & 0.6307  & 0.6607  & 0.6046  & 0.7847  & 0.6540\\
    \hline
    \multirow{3}[2]{*}{C}&\multirow{3}[2]{*}{\checkmark} & \multirow{3}[2]{*}{\checkmark} & \multirow{3}[2]{*}{\ding{55}} & DFEW  & 0.8587  & 0.6738  & 0.7103  & 0.6173  & 0.8237  & 0.6727   \\
          &  &       &       & FERV39K & 0.8340  & 0.6689  & 0.7080  & 0.6146  & 0.8021  & 0.6619   \\
          &  &       &       & MAFW  & 0.8417  & 0.6477  & 0.7357  & 0.6117  & 0.8427  & 0.6695   \\
    \hline   
    \multirow{3}[2]{*}{D}&\multirow{3}[2]{*}{\checkmark}& \multirow{3}[2]{*}{\ding{55}} & \multirow{3}[2]{*}{\checkmark} & DFEW  & \textbf{0.8670} & \textbf{0.6764} & 0.7270  & 0.6198  & 0.8180  & \textbf{0.6810}\\
          & &       &       & FERV39K & 0.8340  & 0.6746  & 0.7017  & 0.6173  & 0.8010  & 0.6774   \\
          &    &       &       & MAFW  & 0.8500  & 0.6612  & \textbf{0.7440} & \textbf{0.6198} & \textbf{0.8427} & 0.6780   \\
    \hline
    \end{tabular}%
    }
  \caption{
  Ablation studies results. \ding{172} H-OTKT module, \ding{173} SRKB with \textit{fixed} \(\sigma_i\), \ding{174} proposed SRKB. }  
  \label{tab:ablation}%
\end{table}%

\noindent\textbf{Influence of H-OTKT module. }To validate the effectiveness of our proposed H-OTKT module, we add the H-OTKT module based on the baseline method (Case \underline{B} in \cref{tab:ablation}). Case \underline{B} contains encoder layer, H-OTKT module and classification layer, it is trained with total loss function Eq. \eqref{eq:loss}. Recall that the H-OTKT captures the optimal relation mapping between VFER classes and deception samples, and performs knowledge transfer. Compared with the results of Case \underline{A}, when DFEW is utilized as source dataset, the ACCs increase on almost all target datasets and modalities, which demonstrates that H-OTKT can facilitate deception detection accuracy by transferring knowledge from the source domain to the target deception domain. Moreover, when source dataset is FERV39K or MAFW, ACCs increase for RLT data with visual modality, while decrease a little for others, because of the different representative abilities in different pre-trained MAE-DFER encoder. Results in Case \underline{B} indicate that H-OTKT lacks robustness for different deception datasets and different modalities.

\noindent\textbf{Influence of SRKB module. }To validate whether the proposed SRKB module is able to improve the robustness of AFFAKT, we add the SRKB module based on the Case \underline{B}. This case corresponds to Case \underline{D} in \cref{tab:ablation}. As we have introduced in the previous section, SRKB is deemed to alleviate the randomness whilst training process, and store and more efficiently use the learned relation mapping by fine-tuning relation mapping during testing (\cref{eq:transportfuse}). From \cref{tab:ablation}, ACCs obviously increase in Case \underline{D} compared with the results in Case \underline{B} on both deception datasets in all modalities. This improvement indicates that the SRKB module is able to boost the performance of AFFAKT. Specifically, when we bring our attention to the results between Case \underline{A}, \underline{B} and \underline{D} on both datasets in visual modality, we are able to discover that SRKB effectively improves the robustness between different datasets. Comparing the results between Case \underline{A}, \underline{B} and \underline{D} in audio and fused modalities, we can conclude that SRKB is able to improve the robustness in different modalities.

\noindent\textbf{Influence of the sample-specific re-weighting strategy in SRKB. } Furthermore, we also validate the effectiveness of our proposed sample-specific re-weighting strategy in SRKB, which is donated as Case \underline{C} in \cref{tab:ablation}. In this case, we \textit{fix} the \(\sigma_i=0.2\) for each sample, which means SRKB would treat all deception samples equally when fine-tuning relation mapping. Compared with results in Case \underline{B}, ACCs on RTL show improvements, but some decreases also appear on DOLOS in audio and fused modalities. Compared with Case \underline{D}, the aforementioned decreases are eliminated. Such resutls illustrate that the sample-specific re-weighting strategy is able to automatically fine-tune the relation mapping to obtain better knowledge transfer guidance when noise in the test deception detection datasets leads to unsatisfactory relation mapping.

\begin{table}[htbp]
\setlength{\tabcolsep}{1.5mm}
  \centering
  \scalebox{0.75}{
    \begin{tabular}{c|cc|cc|cc}
    \hline
    \textbf{Modality} & \multicolumn{2}{c|}{\textbf{Visual}} & \multicolumn{2}{c|}{\textbf{Audio}} & \multicolumn{2}{c}{\textbf{Fused}} \\
    \hline
    \textbf{Dataset} & \textbf{RLT} & \textbf{DOLOS} & \textbf{RLT} & \textbf{DOLOS} & \textbf{RLT} & \textbf{DOLOS} \\
    \hline
    DFEW  & 0.8753   & 0.6751   & 0.7183   & 0.6199   & 0.8177   & 0.6728   \\
    FERV39K & 0.8423   & 0.6711   & 0.7270   & 0.6102   & 0.7927   & 0.6649   \\
    MAFW  & 0.8587   & 0.6697   & 0.7020   & 0.6139   & 0.7927   & 0.6790   \\
    \hline
    \end{tabular}%
    }
  \caption{Results when Former-DFER is employed as pre-trained encoder \(\mathcal{G}\).}
  \label{tab:former-dferres}%
\end{table}%

\noindent\textbf{Influence of pre-trained encoder \(\mathcal{G}\).} In order to evaluate the effect of pre-trained encoder \(\mathcal{G}\), we employ Former-DFER \cite{zhao2021former} as the pre-trained encoder \(\mathcal{G}\) to extract the source features (the process in \cref{fig:overviewmodel} (a)). Comparing the results of \cref{tab:former-dferres} with that of  \cref{tab:comparison}, ACCs on RTL and DOLOS datasets show better performance when the source features are extracted by MAE-DFER. This phenomenon demonstrates that better source feature space structure could facilitate knowledge transformation based on H-OT.

\subsection{Interpretability Studies}

In order to analyze the learned correlation prototype \(\mathbf{B}\), we show the value of \(\mathbf{B}\) in \cref{tab:prototype} when best accuracy is achieved on the two deception datasets by AFFAKT in visual modality. We refer readers to the appendix for the other modalities and their analysis. According to the results shown in \cref{tab:comparison}, the best ACCs are achieved for RLT and DOLOS when the source domain is selected as DFEW in viusal modality. Therefore, the corresponding results for both deception dataset in visual modality is shown in \cref{tab:prototype} when DFEW is employed as the source domain. 
\begin{table}[htb]
  \setlength{\tabcolsep}{1.5mm}
  \centering
  \scalebox{0.65}{
    \begin{tabular}{c|c|ccccccc}
    \hline
     \textbf{Dataset} & \textbf{Category} & \multicolumn{7}{c}{\textbf{Source}} \\
    \hline
     \multirow{4}{*}{\textbf{RLT}} & \textbf{Target} & \textbf{happy} & \textbf{sad} & \textbf{neutral} & \textbf{angry} & \textbf{surprise} & \textbf{disgust} & \textbf{fear} \\
\cline{2-9}            & \textbf{truthtul} & 0.3414 & 0.0999 & 0.1418 & 0.1394 & 0.1903 & 0.0546 & 0.0326 \\
           & \textbf{deceptive} & 0.0318 & 0.4467 & 0.1097 & 0.1190 & 0.1600 & 0.0480 & 0.0847 \\
\cdashline{2-9} & \textbf{DIFF} & \textbf{0.3096} & \textbf{0.3468} & 0.0321 & 0.0204 & 0.0303 & 0.0066 & \textbf{0.0521} \\
\cline{1-9}      \multirow{4}{*}{\textbf{DOLOS}} &       & \textbf{happy} & \textbf{sad} & \textbf{neutral} & \textbf{angry} & \textbf{surprise} & \textbf{disgust} & \textbf{fear}  \\
\cline{2-9} & \textbf{truthtul} & 0.3908 & 0.0936 & 0.1688 & 0.0213 & 0.2343 & 0.0486 & 0.0426  \\
          & \textbf{deceptive} & 0.1324 & 0.3508 & 0.0349 & 0.0923 & 0.2160 & 0.0943 & 0.0793 \\
\cdashline{2-9} & \textbf{DIFF} & \textbf{0.2584} & \textbf{0.2572} & \textbf{0.1339} & 0.0710 & 0.0183 & 0.0457 & 0.0367  \\
    \hline
    \end{tabular}%
    }
  \caption{Learned \(\mathbf{B}\) on RLT and DOLOS in visual modality. \textbf{DIFF} represents the absolute value of the difference between \textit{truthful} and \textit{deceptive}.}
  \label{tab:prototype}%
\end{table}%

There are seven classes (\textit{happy}, \textit{sad}, \textit{neutral}, \textit{angry}, \textit{surprise}, \textit{disgust} and \textit{fear}) in DFEW dataset \cite{jiang2020dfew}  and two classes (\textit{truthful} and \textit{deceptive}) in both deception datasets. Bold represents the largest three absolute difference (DIFF) between being truthful and deceptive across all categories of source domain in \cref{tab:prototype}. 

As demonstrated by \Cref{tab:prototype}, \textit{sad} is significant related to \textit{deceptive} for both RLT and DOLOS, while \textit{happy} is remarkably correlated to \textit{truthful}. Such results are also supported by psychological theory that liars will be less positive and pleasant than truth tellers \cite{depaulo2003cues, zloteanu2020reconsidering, hauch2015computers}. For \textit{neutral} and \textit{fear}, \textit{neutral} has a higher similarity with \textit{truthful} than \textit{deceptive} on both datasets, especially on DOLOS dataset. Besides, \textit{fear} is more correlated with \textit{deceptive} than \textit{truthful} on both datasets. \cite{mathur2020introducing} showed that deceivers in high-stakes situations are likely to associate with \textit{fear}, which is consistent with our results. 

\cref{tab:prototype} shows that AFFAKT can automatically establish proper relations between classes of facial expressions and deception, which helps in leveraging transferred knowledge from facial affective datasets in testing phase.

\section{Conclusions}
\label{sec:limitationsandconclusions}
This paper presents a novel video deception detection method AFFAKT, aiming at addressing the challenge of insufficient high-quality large-scale labeled training datasets. We first develop H-OTKT module to perform knowledge transformation from related facial expression classes to deception samples, which estimates different weights of facial expression to deception samples by H-OT. Moreover, we design a correlation prototype based module SRKB to retain the invariant information within deceptive and truthful samples during training, which maintains the robust relation information between source and target classes. During testing phrase, SRKB integrates the predicted transport plan and the learned correlation prototype using a sample-specific re-weighting technique to leverage transferred knowledge. Extensive experiments have been conducted, showing that our proposed method outperforms other detection methods. 

\appendix

\section{Pseudo-code of AFFAKT}

The pseudo-code of AFFAKT in training phase and testing phase are shown in \cref{alg:trainAFFAKT} and \cref{alg:testAFFAKT}, respectively. The mark $\bigstar$ located in SRKB module in \cref{alg:trainAFFAKT} and \cref{alg:testAFFAKT} is the main difference of forward process between training phase and testing phase. The texts following $\S$ and \# are for explanatory.

\begin{algorithm}[!h]
    \caption{Training Algorithm of AFFAKT}
    \label{alg:trainAFFAKT}
    \renewcommand{\algorithmicrequire}{\textbf{Input:}}
    \renewcommand{\algorithmicensure}{\textbf{Output:}}
    
    \begin{algorithmic}[1]
        \REQUIRE Extracted VFER features $\mathbf{X}^s$; Target video deception detection sample $\mathbf{V}^t$; 
        
        \STATE \# \textit{One Iteration of Training Phase}
        
        \STATE $\S$ \textit{Encode}        
        \STATE $\mathbf{X}^t=\mathcal{E}(\mathbf{V}^t)$
        
        \STATE $\S$ \textit{H-OTKT module}
        \STATE $\mathbf{X}^{t'}=\mathcal{F}_1(\mathbf{X}^t)$
        \FOR{$k \in [1,\dots,L^s]$} 
            \STATE $\mathbf{M}^{low,k}=1-\text{Cosine}(\mathbf{X}^{t'},\mathbf{X}^{s,k})$
            \STATE Compute $\mathbf{T}^{low,k}$ by Eq. (5) \# low-level OT
            \STATE $\mathbf{M}_{:,k}=<\mathbf{T}^{low,k},\mathbf{M}^{low,k}>_{\text{F}}$
        \ENDFOR
        \STATE Compute $\mathbf{T}$ by Eq. (3) \# high-level OT
        \STATE Compute $\mathbf{X}^{trans}$ by Eq. (6)
        \STATE Compute $\mathbf{X}^{fused}$ by Eq. (7)     
        
        \STATE $\S$ \textit{SRKB module}
        \STATE Update $\mathbf{B}$ by Eq. (12)  $\bigstar$

        \STATE $\S$ \textit{Classification, Loss Calculation and Backward}
        \STATE Compute $\hat{\boldsymbol{y}}$ by Eq. (8)
        \STATE Calculate $\mathcal{L}$ by Eq. (9), Eq. (10) and Eq. (11)
        \STATE Backpropagation $Backward(\mathcal{L})$
    \end{algorithmic}
\end{algorithm}
\begin{algorithm}[!h]
    \caption{Testing Algorithm of AFFAKT}
    \label{alg:testAFFAKT}
    \renewcommand{\algorithmicrequire}{\textbf{Input:}}
    \renewcommand{\algorithmicensure}{\textbf{Output:}}
    
    \begin{algorithmic}[1]
        \REQUIRE Extracted VFER features $\mathbf{X}^s$; Target video deception detection sample $\mathbf{V}^t$; 

        \STATE \# \textit{One Iteration of Testing Phase}
        \STATE $\S$ \textit{Encode}        
        \STATE $\mathbf{X}^t=\mathcal{E}(\mathbf{V}^t)$
        
        \STATE $\S$ \textit{H-OTKT module}
        \STATE $\mathbf{X}^{t'}=\mathcal{F}_1(\mathbf{X}^t)$
        \FOR{$k \in [1,\dots,L^s]$} 
            \STATE $\mathbf{M}^{low,k}=1-\text{Cosine}(\mathbf{X}^{t'},\mathbf{X}^{s,k})$
            \STATE Compute $\mathbf{T}^{low,k}$ by Eq. (5) \# low-level OT
            \STATE $\mathbf{M}_{:,k}=<\mathbf{T}^{low,k},\mathbf{M}^{low,k}>_{\text{F}}$
        \ENDFOR
        \STATE Compute $\mathbf{T}$ by Eq. (3) \# high-level OT
        \STATE Compute $\mathbf{X}^{trans}$ by Eq. (6)
        \STATE Compute $\mathbf{X}^{fused}$ by Eq. (7)     
        
        \STATE $\S$ \textit{SRKB module}
        \STATE Re-weight $\hat{\mathbf{T}}$ by Eq. (13) $\bigstar$

        \STATE $\S$ \textit{Classification}
        \STATE Compute $\hat{\boldsymbol{y}}$ by Eq. (8)
    \end{algorithmic}
\end{algorithm}

\section{Datasets and Experimental Settings}

\subsection{Datasets}
\noindent\textbf{Video Deception Detection Datasets. }We conduct the experiments on two most widely used datasets in video deception detection task,  Real-Life Trial (RLT) dataset \cite{perez2015deception} and DOLOS dataset \cite{guo2023audio}. RLT dataset contains 121 video samples (61 deceptive samples and 60 truthful samples) with an average length of 28.0 seconds from real life court trial recordings. DOLOS dataset was recently generated from gameshow videos \cite{guo2023audio}, it contains 1675 video samples (899 deceptive and 776 truthful) ranging from 2 to 19 seconds.

\noindent\textbf{Video Facial Expression Recognition Datasets. }Three \textit{in-the-wild} VFER datasets (DFEW \cite{jiang2020dfew}, FERV39K \cite{wang2022ferv39k} and MAFW \cite{liu2022mafw}) are employed in our experiments. They contain 16372, 38935 and 10045 samples with 7, 7, and 11 expression categories, respectively. For DFEW and MAFW, we only use 11697 and 9172 single-labeled clips, respectively.

\subsection{Experimental Settings}

Our experiments are conducted with Ubuntu 20.04, Python 3.11 with PyTorch 2.0.0 \cite{paszke2017automatic} on one RTX 3090 24GB and one A800 80GB GPUs.
16 video frames are sampled uniformly from each sample in target deception dataset, and then resized to \(224\times 224\). Pre-trained encoders VideoMAE \cite{tong2022videomae} and W2V2  \cite{baevski2020wav2vec} are exploited as backbones to encode visual and audio data separately, the output of backbone performs as the input of H-OTKT. For VFER dataset, pre-trained encoder MAE-DFER, which gained the most compact and separable embedding space \cite{sun2023mae}, is utilized to generate the visual representations, where Former-DFER \cite{zhao2021former} is also employed in ablation studies. 

We use Adam optimizer with learning rate as $0.00001$. The batch size is $4$ in training phase and $2$ in testing phase. Our model is trained for $20$ epochs with $5$-fold cross validation. \(\alpha\) and \(\eta\) are set to $0.95$ and $0.01$, respectively. For RLT, 
\(\xi\) is set as $0.5$ for audio modality and $0.2$ for visual and fused modalities. For DOLOS, \(\xi\) is set to \(0.2\) for all modalities. \(\nu\) is $0.05$ for visual and fused modalities in RLT, and \(\nu\) is $0.1$ for audio modality in RLT and all modalities in DOLOS. Accuracy (ACC), F1 score and area under the curve (AUC) are adopted for evaluation. Particularly, PECL(visual) \cite{guo2023audio} is employed in the toy experiment to obtain Fig. (1), which is trained for $100$ epochs on a single fold of DOLOS.

\section{Detailed Description about Comparison Methods}
\label{sec:appendixdetaileddescription}
In this section, we will give more detailed descriptions about the comparison methods.
\subsection{Traditional Machine Learning based Deception Detection Methods}
Firstly, we would like to introduce the statistical features that are used in our experiments.
\begin{itemize}
    \item Visual: OpenFace \cite{baltrusaitis2018openface} is an open source tool for extracting facial statistical features, such as landmarks, action units. Some of the action units also show high association with deception \cite{csen2020multimodal}. Following the previous researches \cite{mathur2020introducing, Krishnamurthy2018ADL, avola2019automatic, yang2021multimodal}, we also employ OpenFace as our visual feature extractor to obtain visual statistical features.
    \item Audio: Mel-scale Frequency Cepstral Coefficients (MFCC) \cite{9955539} and OpenSMILE \cite{eyben2010opensmile} are two mostly used acoustic statistical features for detecting deception. 
\end{itemize}

Not that the features extracted by OpenFace are frame-wise, since different video clips may contains different number of frames, we normalize the dimension of one video clip by OpenMM \cite{morales2017openmm}, which calculates the 11 statistical functionals for each feature at view label \cite{rill2019high}.

For classification, we employ SVM, Decision Tree Random Forest, and AdaBoost as our classifier. The statistical features are fed to each classifier to perform classification.

\subsection{Deep Learning based Deception Detection Methods}

In this paper, we make comparisons with several deep learning methods. These methods can be separated by their backbone structure: Long Short-Term Memory (LSTM) \cite{graves2012long} based, ResNet \cite{he2016deep} based, and Transformer \cite{vaswani2017attention} based.

\begin{itemize}
    \item LSTM based: LSTM is known as the sequence encoder, which is able to capture the contextual information of a given sequence. In this case, LSTM is employed to handle the contextual information aggregation at temporal dimension. Several researches adopt LSTM as the temporal encoder to obtain the temporal information \cite{mathur2020introducing,Krishnamurthy2018ADL,guo2023audio}.
    \item ResNet based: ResNet is a common image encoder, which is built upon convolutional neural networks. In these works \cite{9446553, Ding_2019_CVPR, Krishnamurthy2018ADL,guo2023audio}, they use ResNet to automatically extract the visual features instead of using OpenFace or  other manual approaches. In our experiments, ResNet with 18 layers (ResNet18) is employed to extract the visual features of each video frames.
    \item Transformer based: With the great success of Transformer \cite{vaswani2017attention}, encoders with more parameters based on Transformer architecture have been proposed to encode video clips or audio sequences automatically with rich semantic information. W2V2 \cite{baevski2020wav2vec} is typical audio encoder based on Transformer architecture, and VideoMAE \cite{tong2022videomae} is able to directly encode the given video clip to a fixed length vector. 
\end{itemize}

In our experiment, we make comparisons with the following researches.

\begin{itemize}
    \item ResNet18 + LSTM \cite{9446553, Ding_2019_CVPR, Krishnamurthy2018ADL,guo2023audio}: In these methods, ResNet18 was adopted to extract video frame features of a video. Then sequential information of all frame features was formulated by an LSTM. Then an MLP performed classification using the last output feature of the sequence.
    \item W2V2 + MLP \cite{guo2023audio,9446553,Krishnamurthy2018ADL}: In these methods, W2V2 model was used to extract audio features. Then an MLP was used to make classification.
    \item ResNet18 \(\oplus\) OpenSMILE \cite{gogate2017deep,Krishnamurthy2018ADL,guo2023audio}: These methods took both visual and audio modalities into account, and performed late fusion from each single modality branch.
\end{itemize}

\subsection{Transfer Learning based Methods}

There have been a small number of researches that tried to transfer knowledge from other related dataset to enhance the detection performance with deep learning based methods. Therefore, we adapt several common kinds of transfer learning strategy to the deception detection task. 

\begin{itemize}
    \item Optimal Transport based: Free Lunch \cite{yang2021free} achieved knowledge transfer by estimating the weight of each base class and perform distribution calibration with the statistics of base classes, which directly used the distance of class average feature and support feature as the measurement. Similar to FreeLunch, ADC \cite{guo2022adaptive} also aimed to transfer knowledge via quantifying the weight of each source class and target sample and perform distribution calibration. The optimal transport plan represents the importance (or correlation) between the base classes and the novel samples. In our case, the relation between deception sample and each facial expression class is estimated by these two methods in the forward process, which play roles same as the H-OTKT and SRKB modules.
    \item Pre-train \& Fine-tune based: The transfer learning methods of this kind are more likely to be adopted in large models, such as PECL \cite{guo2023audio}. It tried to transfer knowledge from the pre-trained dataset and checkpoint to the target dataset.
    \item Knowledge Distillation based: Knowledge distillation is also known as a typical transfer learning method. In \cite{gu2023class}, knowledge distillation was used for discover novel class samples given a model pre-trained on a source dataset. The key idea of \cite{gu2023class} is to distill knowledge according to the class realtion.
\end{itemize}

\section{The Standard Deviation Report Between Folds}

The average value and the standard deviation between different folds are shown in \cref{tab:comparison}. Beside the analysis in the main text, the results in \cref{tab:comparison} show that AFFAKT is more robust, since the evaluate metric between different folds have smaller standard deviation value.

\begin{table*}[!tb]
  \centering
  \begin{subtable}{\linewidth} 
  \centering
  \scalebox{0.7}{
    \begin{tabular}{c|c|ccc|ccc}
    \hline
    \multicolumn{2}{c|}{\textbf{Target}} & \multicolumn{3}{c|}{\textbf{RLT}} & \multicolumn{3}{c}{\textbf{DOLOS}}\\
    \hline
    \textbf{Method} & \textbf{Source Dataset} & \textbf{F1 score} & \textbf{ACC} & \textbf{AUC} & \textbf{F1 score} & \textbf{ACC} & \textbf{AUC} \\
    \hline
    OpenFace + SVM  & -     & {0.2253$\pm$0.2605} & {0.5293$\pm$0.0361} & {0.5571$\pm$0.0470} & {0.6975$\pm$0.0010} & {0.5355$\pm$0.0012} & {0.5430$\pm$0.0160}\\
    OpenFace + Decision Tree  & -     & {0.5553$\pm$0.1157} & {0.5303$\pm$0.1048} & {0.5303$\pm$0.1048} & {0.5358$\pm$0.0303} & {0.5058$\pm$0.0262} & {0.5058$\pm$0.0262} \\
    OpenFace + Random Forest  & -     & {0.6033$\pm$0.0867} & {0.6033$\pm$0.0559} & {0.5997$\pm$0.0574} & {0.6175$\pm$0.0193} & {0.5367$\pm$0.0227} & {0.5466$\pm$0.0272} \\
    OpenFace + AdaBoost & -     & {0.5199$\pm$0.1523} & {0.5303$\pm$0.0980} & {0.5766$\pm$0.1070} & {0.5536$\pm$0.0251} & {0.5057$\pm$0.0329} & {0.5035$\pm$0.0357}\\
    \hline
    AU + SVM  & -     & {0.4562$\pm$0.0723} & {0.5043$\pm$0.0726} & {0.4670$\pm$0.0970} & {0.6813$\pm$0.0194} & {0.5276$\pm$0.0126} & {0.5242$\pm$0.0089}\\
    AU + Decision Tree  & -     & {0.4466$\pm$0.1577} & {0.4643$\pm$0.1167} & {0.4643$\pm$0.1167} & {0.5453$\pm$0.0172} & {0.5173$\pm$0.0137} & {0.5173$\pm$0.0137} \\
    AU + Random Forest & -     & {0.5534$\pm$0.0792} & {0.5463$\pm$0.0810} & {0.5330$\pm$0.0766} & {0.5808$\pm$0.0183} & {0.5045$\pm$0.0256} & {0.5157$\pm$0.0230} \\
    AU + AdaBoost  & -     & {0.5130$\pm$0.0530} & {0.4877$\pm$0.0612} & {0.4835$\pm$0.0833} & {0.5295$\pm$0.0302} & {0.4876$\pm$0.0185} & {0.4735$\pm$0.0264} \\
    \hline OpenFace + LSTM & -     & {0.5241$\pm$0.0995} & {0.5623$\pm$0.0834} & {0.5952$\pm$0.1164} & {0.5928$\pm$0.0342} & {0.5628$\pm$0.0164} & {0.5854$\pm$0.0152}\\
    AU + LSTM  & -     & {0.4888$\pm$0.0472} & {0.6197$\pm$0.0419} & {0.6760$\pm$0.0442} & {0.6343$\pm$0.0084} & {0.5646$\pm$0.0137} & {0.5868$\pm$0.0098} \\
    ResNet18 + LSTM & -     & {0.4996$\pm$0.1391} & {0.6117$\pm$0.0718} & {0.6387$\pm$0.0928} & {0.6415$\pm$0.0124} & {0.5972$\pm$0.0087} & {0.5668$\pm$0.0136}\\
    \hline
    PECL(only visual) & -     & {0.5880$\pm$0.1018} & {0.6528$\pm$0.0040} & {0.6734$\pm$0.0508} & {0.7010$\pm$0.0213} & {0.6387$\pm$0.0139} & {0.6770$\pm$0.0099} \\
    \hline
    \multirow{3}[2]{*}{FreeLunch } & DFEW  & {0.7612$\pm$0.1207} & {0.8090$\pm$0.0782} & {0.8712$\pm$0.0782} & {0.6961$\pm$0.0147} & {0.6222$\pm$0.0221} & {0.6444$\pm$0.0221} \\
          & FERV39K & {0.7536$\pm$0.1323} & {0.8173$\pm$0.0781} & {0.8677$\pm$0.0781} & {0.6831$\pm$0.0036} & {0.6228$\pm$0.0179} & {0.6456$\pm$0.0179} \\
          & MAFW  & {0.7663$\pm$0.1135} & {0.8173$\pm$0.0781} & {0.8633$\pm$0.0781} & {0.6695$\pm$0.0183} & {0.6155$\pm$0.0197} & {0.6459$\pm$0.0197} \\
    \hline
    \multirow{3}[2]{*}{ADC } & DFEW  & {0.7793$\pm$0.1218} & {0.8173$\pm$0.0942} & {0.8674$\pm$0.0943} & {0.6880$\pm$0.0163} & {0.6716$\pm$0.0157} & {0.7206$\pm$0.0157} \\
          & FERV39K & {0.7667$\pm$0.1406} & {0.8173$\pm$0.0943} & {0.8677$\pm$0.0943} & {0.6830$\pm$0.0198} & {0.6693$\pm$0.0114} & {0.7156$\pm$0.0114} \\
          & MAFW  & {0.7667$\pm$0.1406} & {0.8173$\pm$0.0943} & {0.8664$\pm$0.0942} & {0.6938$\pm$0.0238} & {0.6684$\pm$0.0170} & {0.7180$\pm$0.0170} \\
    \hline
    \multirow{3}[2]{*}{Cr-KD-NCD } & DFEW  & {0.6957$\pm$0.1342} & {0.7200$\pm$0.0869} & {0.6928$\pm$0.0865} & {0.5850$\pm$0.0175} & {0.6091$\pm$0.0186} & {0.6013$\pm$0.0202} \\
          & FERV39K & {0.7805$\pm$0.1136} & {0.7200$\pm$0.0756} & {0.6464$\pm$0.0803} & {0.6720$\pm$0.0120} & {0.5879$\pm$0.0135} & {0.5363$\pm$0.0233} \\
          & MAFW  & {0.7778$\pm$0.1029} & {0.6800$\pm$0.0924} & {0.6368$\pm$0.0926} & {0.7056$\pm$0.0196} & {0.5697$\pm$0.0145} & {0.5427$\pm$0.0129} \\
    \hline
    \multirow{3}[2]{*}{AFFAKT (ours)} & DFEW  & \textbf{{0.8760$\pm$0.0516}} & \textbf{{0.8670$\pm$0.0558}} & \textbf{{0.8789$\pm$0.0516}} & {0.7054$\pm$0.0196} & \textbf{{0.6764$\pm$0.0199}} & \textbf{{0.7212$\pm$0.0292}} \\
          & FERV39K & {0.8277$\pm$0.1041} & {0.8340$\pm$0.0753} & {0.8415$\pm$0.0970} & \textbf{0.7102$\pm$0.0233} & {0.6746$\pm$0.0197} & {0.7203$\pm$0.0287} \\
          & MAFW  & {0.8524$\pm$0.0772} & {0.8500$\pm$0.0773} & {0.8625$\pm$0.0722} & 0.6948$\pm$0.0299 & {0.6612$\pm$0.0233} & {0.6970$\pm$0.0236}\\
    \hline
    \end{tabular}%
    }
    \subcaption{Results with visual modality. }
    \label{tab:comparisonvisual}
    \end{subtable}
    \begin{subtable}{\linewidth} 
    \centering
        \scalebox{0.7}{
        \begin{tabular}{c|c|ccc|ccc}
        \hline
        \multicolumn{2}{c|}{\textbf{Target Dataset}} & \multicolumn{3}{c|}{\textbf{RLT}} & \multicolumn{3}{c}{\textbf{DOLOS}} \\
        \hline
        \textbf{Method} & \textbf{Source Dataset} & \textbf{F1 score} & \textbf{ACC} & \textbf{AUC} & \textbf{F1 score} & \textbf{ACC} & \textbf{AUC} \\
        \hline
        MFCC + MLP  & -     & 0.5226$\pm$0.2911 & 0.6367$\pm$0.1263 & 0.7030$\pm$0.0502 & 0.5963$\pm$0.0757 & 0.5810$\pm$0.0232 & 0.6134$\pm$0.0279 \\
        OpenSMILE + MLP & -     & 0.6885$\pm$0.1275 & 0.6597$\pm$0.1121 & 0.5926$\pm$0.0916 & 0.6867$\pm$0.0128 & 0.5537$\pm$0.0095 & 0.5325$\pm$0.0091 \\
        W2V2 + MLP  & -     & 0.6117$\pm$0.0810 & 0.6780$\pm$0.0266 & 0.6106$\pm$0.0631 & 0.4383$\pm$0.0333 & 0.5421$\pm$0.0115 & 0.5369$\pm$0.0120 \\
        \hline
        PECL(only audio) & -     & 0.7121$\pm$0.0748 & 0.7100$\pm$0.0718 & 0.6962$\pm$0.0796 & 0.6777$\pm$0.0364 & 0.6119$\pm$0.0200 & 0.6281$\pm$0.0155 \\
        \hline
        \multirow{3}[1]{*}{FreeLunch} & DFEW  & 0.6396$\pm$0.0864 & 0.6767$\pm$0.0832 & 0.6869$\pm$0.0832 & 0.6437$\pm$0.0425 & 0.5864$\pm$0.0213 & 0.6157$\pm$0.0213 \\
          & FERV39K & 0.6432$\pm$0.0989 & 0.6850$\pm$0.0704 & 0.6944$\pm$0.0704 & 0.6589$\pm$0.0334 & 0.5979$\pm$0.0203 & 0.6194$\pm$0.0203 \\
          & MAFW  & 0.6402$\pm$0.0922 & 0.6767$\pm$0.0744 & 0.6885$\pm$0.0744 & 0.6490$\pm$0.0396 & 0.5991$\pm$0.0188 & 0.6196$\pm$0.0188 \\
        \hline
        \multirow{3}[1]{*}{ADC} & DFEW  & 0.6402$\pm$0.0922 & 0.6767$\pm$0.0744 & 0.6858$\pm$0.0744 & 0.6196$\pm$0.0814 & 0.6058$\pm$0.0135 & 0.6040$\pm$0.0135 \\
          & FERV39K & 0.6402$\pm$0.0922 & 0.6767$\pm$0.0744 & 0.6858$\pm$0.0744 & 0.6165$\pm$0.0854 & 0.6052$\pm$0.0116 & 0.6039$\pm$0.0116 \\
          & MAFW  & 0.6272$\pm$0.1133 & 0.6767$\pm$0.0744 & 0.6842$\pm$0.0744 & 0.6129$\pm$0.0835 & 0.6046$\pm$0.0125 & 0.6039$\pm$0.0125 \\
        \hline
        \multirow{3}[0]{*}{AFFAKT (ours)} & DFEW  & 0.7267$\pm$0.0242 & 0.7270$\pm$0.0350 & 0.7218$\pm$0.0631 & 0.6822$\pm$0.0322 & 0.6198$\pm$0.0165 & \textbf{0.6391$\pm$0.0184} \\
              & FERV39K & \textbf{0.7316$\pm$0.0493} & 0.7017$\pm$0.1016 & 0.6917$\pm$0.0956 & \textbf{0.6982$\pm$0.0210} & 0.6173$\pm$0.0181 & 0.6385$\pm$0.0273 \\
              & MAFW  & 0.7266$\pm$0.0867 & \textbf{0.7440$\pm$0.0707} & \textbf{0.7396$\pm$0.0768} & 0.6736$\pm$0.0272 & \textbf{0.6198$\pm$0.0076} & 0.6387$\pm$0.0080 \\
            \hline
        \end{tabular}%
        }
        \label{tab:comparisonaudio}%
        \subcaption{Results with audio modality. }
    \end{subtable}
    \begin{subtable}{\linewidth} 
    \centering
    \scalebox{0.7}{
    \begin{tabular}{c|c|ccc|ccc}
    \hline
    \multicolumn{2}{c|}{\textbf{Target Dataset}} & \multicolumn{3}{c|}{\textbf{RLT}} & \multicolumn{3}{c}{\textbf{DOLOS}} \\
    \hline
    \textbf{Method} & \textbf{Source Dataset} & \textbf{F1 score} & \textbf{ACC} & \textbf{AUC} & \textbf{F1 score} & \textbf{ACC} & \textbf{AUC} \\
    \hline
    OpenFace \(\oplus\) OpenSMILE & -     & 0.6895$\pm$0.0463 & 0.6781$\pm$0.0752 & 0.6212$\pm$0.0671 & 0.6124$\pm$0.0354 & 0.5986$\pm$0.0153 & 0.5863$\pm$0.0136 \\
    ResNet18 \(\oplus\) OpenSMILE  & -     & 0.6283$\pm$0.0498 & 0.6853$\pm$0.0627 & 0.6598$\pm$0.0763 & 0.5863$\pm$0.0263 & 0.6152$\pm$0.0175 & 0.6485$\pm$0.0121 \\
    \hline
    PECL & -     & 0.7102$\pm$0.0215 & 0.6939$\pm$0.0488 & 0.7424$\pm$0.0569 & 0.7084$\pm$0.0142 & 0.6597$\pm$0.0114 & 0.6353$\pm$0.0108 \\
    \hline
    \multirow{3}[1]{*}{FreeLunch} & DFEW  & 0.7473$\pm$0.0826 & 0.8010$\pm$0.0859 & 0.8497$\pm$0.0859 & 0.6686$\pm$0.0373 & 0.6258$\pm$0.0119 & 0.6628$\pm$0.0119 \\
          & FERV39K & 0.7460$\pm$0.0945 & 0.8010$\pm$0.0859 & 0.8504$\pm$0.0859 & 0.6504$\pm$0.0695 & 0.6204$\pm$0.0138 & 0.6574$\pm$0.0138 \\
          & MAFW  & 0.7695$\pm$0.0799 & 0.8093$\pm$0.0782 & 0.8547$\pm$0.0781 & 0.6807$\pm$0.0251 & 0.6289$\pm$0.0060 & 0.6669$\pm$0.0060 \\
    \hline
    \multirow{3}[1]{*}{ADC} & DFEW  & 0.7493$\pm$0.0703 & 0.8093$\pm$0.0782 & 0.8446$\pm$0.0782 & 0.6997$\pm$0.0010 & 0.6746$\pm$0.0126 & 0.7307$\pm$0.0115 \\
          & FERV39K & 0.7493$\pm$0.0703 & 0.8093$\pm$0.0782 & 0.8435$\pm$0.0782 & 0.6819$\pm$0.0104 & 0.6729$\pm$0.0183 & 0.7295$\pm$0.0106 \\
          & MAFW  & 0.7493$\pm$0.0703 & 0.8010$\pm$0.0817 & 0.8411$\pm$0.0817 & 0.6976$\pm$0.0124 & 0.6741$\pm$0.0185 & 0.7274$\pm$0.0091 \\
    \hline
    \multirow{3}[0]{*}{AFFAKT (ours)} & DFEW  & 0.8162$\pm$0.0654 & 0.8180$\pm$0.0628 & 0.8381$\pm$0.0737 & 0.7073$\pm$0.0303 & \textbf{0.6810$\pm$0.0140} & 0.7226$\pm$0.0192 \\
          & FERV39K & 0.7946$\pm$0.0923 & 0.8010$\pm$0.0817 & 0.8357$\pm$0.0798 & \textbf{0.7149$\pm$0.0099} & 0.6774$\pm$0.0038 & \textbf{0.7289$\pm$0.0092} \\
          & MAFW  & \textbf{0.8412$\pm$0.0848} & \textbf{0.8427$\pm$0.0768} & \textbf{0.8563$\pm$0.0688} & 0.7111$\pm$0.0146 & 0.6780$\pm$0.0148 & 0.7181$\pm$0.0330 \\
          \hline
            \end{tabular}%
        }
        \label{tab:comparisonva}%
        \subcaption{Results with fused modalities.
        }
    \end{subtable}
  \caption{Comparison results on Real-Life Trial dataset (RLT) \cite{perez2015deception} and DOLOS \cite{guo2023audio} dataset with F1 score, ACC and AUC metrics. Both mean and standard deviation are reported (mean\(\pm\)std). }  
  \label{tab:comparison}%
\end{table*}%

\begin{table*}[tb]
    \renewcommand\arraystretch{1.3}
  \centering
  \scalebox{0.7}{
    \begin{tabular}{c|c|c|ccccccccccc}
    \hline
    \textbf{Modality} & \textbf{Dataset} & \textbf{Category} & \multicolumn{11}{c}{\textbf{Source Domain}} \\
    \hline
    \multirow{8}{*}{\textbf{Visual}} & \multirow{4}{*}{\textbf{RLT}} & \textbf{Target Domain} & \textbf{happy} & \textbf{sad} & \textbf{neutral} & \textbf{angry} & \textbf{surprise} & \textbf{disgust} & \textbf{fear} & \textbf{-} & \textbf{-} & \textbf{-} & \textbf{-} \\
\cline{3-14}          &       & \textbf{truthtul} & 0.3414 & 0.0999 & 0.1418 & 0.1394 & 0.1903 & 0.0546 & 0.0326 &  -     &    -   &  -     & - \\
          &       & \textbf{deceptive} & 0.0318 & 0.4467 & 0.1097 & 0.1190 & 0.1600 & 0.0480 & 0.0847 &    -   &   -    &  -     &-  \\
\cdashline{3-14}          &       & \textbf{DIFF} & \textbf{0.3096} & \textbf{0.3468} & 0.0321 & 0.0204 & 0.0303 & 0.0066 & \textbf{0.0521} & - & - & - & - \\
\cline{2-14}          & \multirow{4}{*}{\textbf{DOLOS}} &       & \textbf{happy} & \textbf{sad} & \textbf{neutral} & \textbf{angry} & \textbf{surprise} & \textbf{disgust} & \textbf{fear} & -     & -     & -     & - \\
\cline{3-14}          &       & \textbf{truthtul} & 0.3908 & 0.0936 & 0.1688 & 0.0213 & 0.2343 & 0.0486 & 0.0426 & -     & -     & -     & - \\
          &       & \textbf{deceptive} & 0.1324 & 0.3508 & 0.0349 & 0.0923 & 0.2160 & 0.0943 & 0.0793 & -     & -     & -     & - \\
\cdashline{3-14}          &       & \textbf{DIFF} & \textbf{0.2584} & \textbf{0.2572} & \textbf{0.1339} & 0.0710 & 0.0183 & 0.0457 & 0.0367 & -     & -     & -     & - \\
    \hline
    \multirow{8}{*}{\textbf{Audio}} & \multirow{4}{*}{\textbf{RLT}} & \textbf{Target Domain} & \textbf{anger} & \textbf{disgust} & \textbf{fear} & \textbf{happiness} & \textbf{neutral} & \textbf{sad} & \textbf{surprise} & \textbf{contempt} & \textbf{anxiety} & \textbf{helpless} & \textbf{disappoint} \\
\cline{3-14}          &       & \textbf{truthtul} & 0.0946 & 0.0150 & 0.1328 & 0.3109 & 0.2097 & 0.1246 & 0.0362 & 0.0219 & 0.0040 & 0.0238 & 0.0266 \\
          &       & \textbf{deceptive} & 0.0654 & 0.1072 & 0.1967 & 0.0801 & 0.1019 & 0.2485 & 0.0997 & 0.0392 & 0.0236 & 0.0200 & 0.0177\\
\cdashline{3-14}          &       & \textbf{DIFF} & 0.0292 & 0.0922 & 0.0639 & \textbf{0.2308} & \textbf{0.1078} & \textbf{0.1239} & 0.0635 & 0.0173 & 0.0196 & 0.0038 & 0.0089 \\
\cline{2-14}          & \multirow{4}{*}{\textbf{DOLOS}} &       & \textbf{anger} & \textbf{disgust} & \textbf{fear} & \textbf{happiness} & \textbf{neutral} & \textbf{sad} & \textbf{surprise} & \textbf{contempt} & \textbf{anxiety} & \textbf{helpless} & \textbf{disappoint} \\
\cline{3-14}          &       & \textbf{truthtul} & 0.0864 & 0.0655 & 0.1466 & 0.1903 & 0.0275 & 0.0943 & 0.1595 & 0.0612 & 0.0188 & 0.0941 & 0.0558\\
          &       & \textbf{deceptive} & 0.2146 & 0.1275 & 0.1122 & 0.1541 & 0.0256 & 0.0686 & 0.0893 & 0.0682 & 0.0404 & 0.0621 & 0.0373 \\
\cdashline{3-14}          &       & \textbf{DIFF} & \textbf{0.1282} & 0.0620 & 0.0226 & \textbf{0.0932} & 0.0419 & 0.0143 & \textbf{0.0702} & 0.0070 & 0.0216 & 0.0320 & 0.0185 \\
    \hline
    \multirow{8}{*}{\textbf{Fused}} & \multirow{4}{*}{\textbf{RLT}} & \textbf{Target Domain} & \textbf{anger} & \textbf{disgust} & \textbf{fear} & \textbf{happiness} & \textbf{neutral} & \textbf{sad} & \textbf{surprise} & \textbf{contempt} & \textbf{anxiety} & \textbf{helpless} & \textbf{disappoint} \\
\cline{3-14}          &       & \textbf{truthtul} & 0.1074 & 0.1514 & 0.0316 & 0.3407 & 0.1100 & 0.1266 & 0.0396 & 0.0250 & 0.0168 & 0.0160 & 0.0348 \\
          &       & \textbf{deceptive} & 0.0824 & 0.0439 & 0.2351 & 0.0719 & 0.2357 & 0.1262 & 0.0552 & 0.0652 & 0.0398 & 0.0249 & 0.0200 \\
\cdashline{3-14}          &       & \textbf{DIFF} & 0.0250 & 0.1075 & \textbf{0.2035} & \textbf{0.2688} & \textbf{0.1257} & 0.0004 & 0.0156 & 0.0402 & 0.0230 & 0.0089 & 0.0148 \\
\cline{2-14}          & \multirow{4}{*}{\textbf{DOLOS}} &       & \textbf{happy} & \textbf{sad} & \textbf{neutral} & \textbf{angry} & \textbf{surprise} & \textbf{disgust} & \textbf{fear} & -     & -     & -     & - \\
\cline{3-14}          &       & \textbf{truthtul} & 0.3101 & 0.0881 & 0.2577 & 0.0923 & 0.1598 & 0.0448 & 0.0471 & -     & -     & -     & - \\
          &       & \textbf{deceptive} & 0.1942 & 0.1329 & 0.1704 & 0.1490 & 0.2056 & 0.0148 & 0.1331 & -     & -     & -     & - \\
\cdashline{3-14}          &       & \textbf{DIFF} & \textbf{0.1159} & 0.0448 & \textbf{0.0873} & 0.0567 & 0.0458 & 0.0300 & \textbf{0.0860} & -     & -     & -     & - \\
    \hline
    \end{tabular}%
    }
  \caption{Learned correlation prototype \(\mathbf{B}\) on RLT and DOLOS. \textbf{DIFF} represents the absolute value of the difference between \textit{truthful} and \textit{deceptive}.}
  \label{tab:prototype}%
\end{table*}%

\section{Sensitive Analyses}
\label{sec:sensitive}
\begin{figure}[!b]
  \centering
    \begin{subfigure}{0.45\linewidth}
        \centering
        \includegraphics[width=1\linewidth]{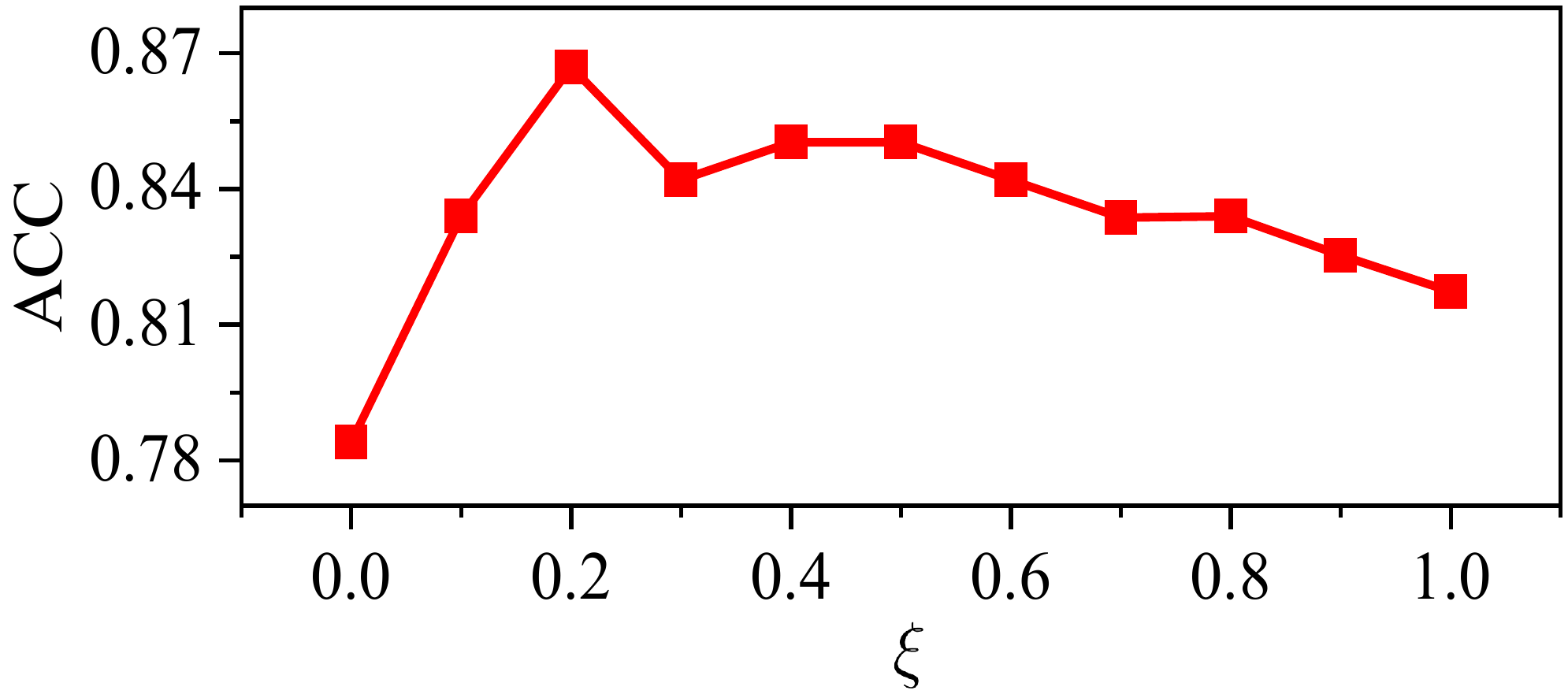}
        \caption{ACC with different \(\xi\)}
        \label{fig:sensitive1}
    \end{subfigure}
    \begin{subfigure}{0.45\linewidth}
        \centering
        \includegraphics[width=1\linewidth]{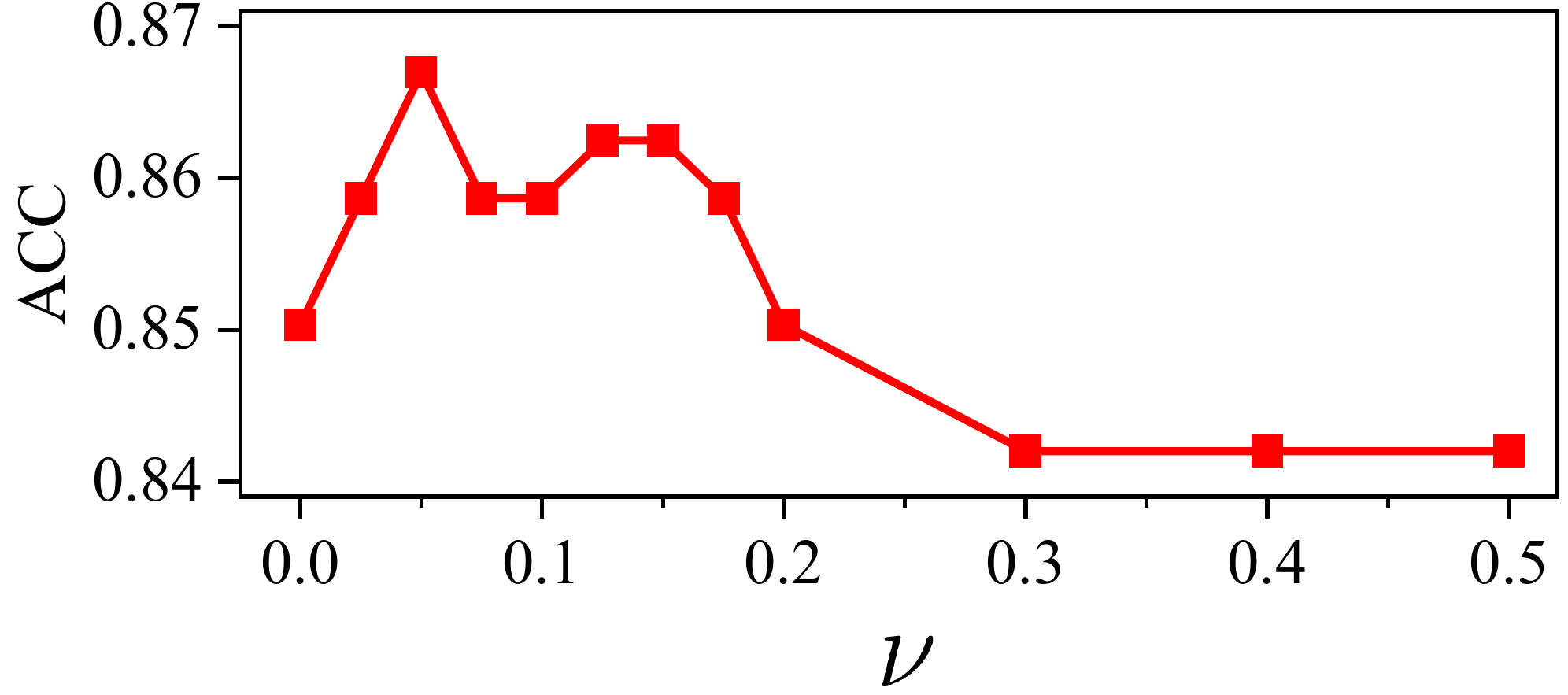}
        \caption{ACC with different \(\nu\)}
        \label{fig:sensitive2}
    \end{subfigure}
    \begin{subfigure}{0.45\linewidth}
        \centering
        \includegraphics[width=1\linewidth]{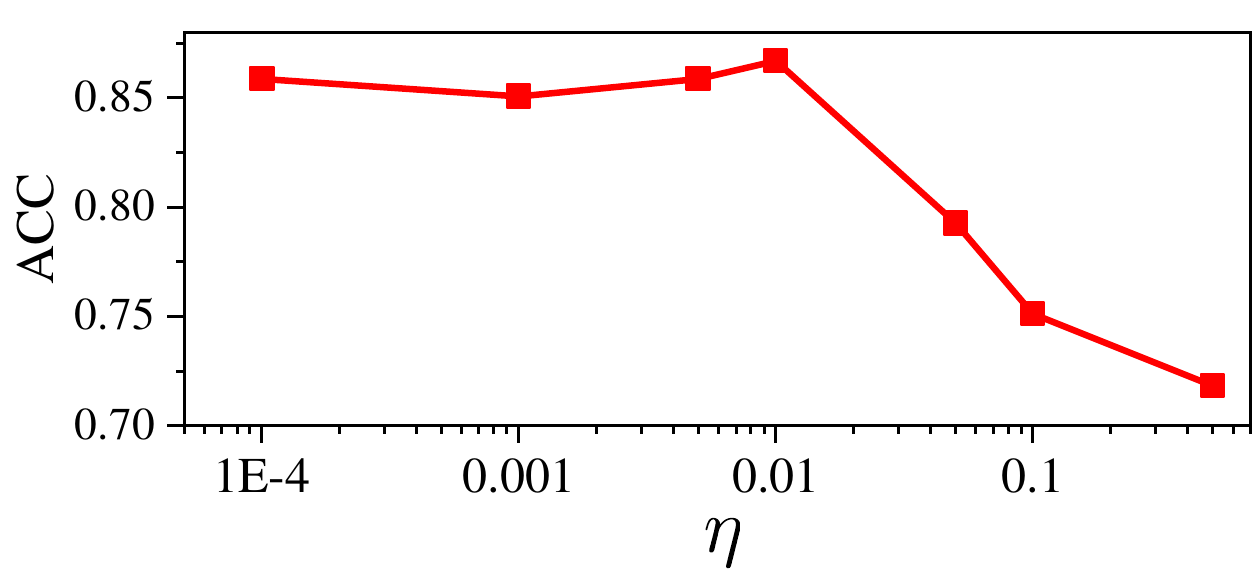}
        \caption{ACC with different \(\eta\)}
        \label{fig:sensitive3}
    \end{subfigure}
    \begin{subfigure}{0.45\linewidth}
        \centering
        \includegraphics[width=1\linewidth]{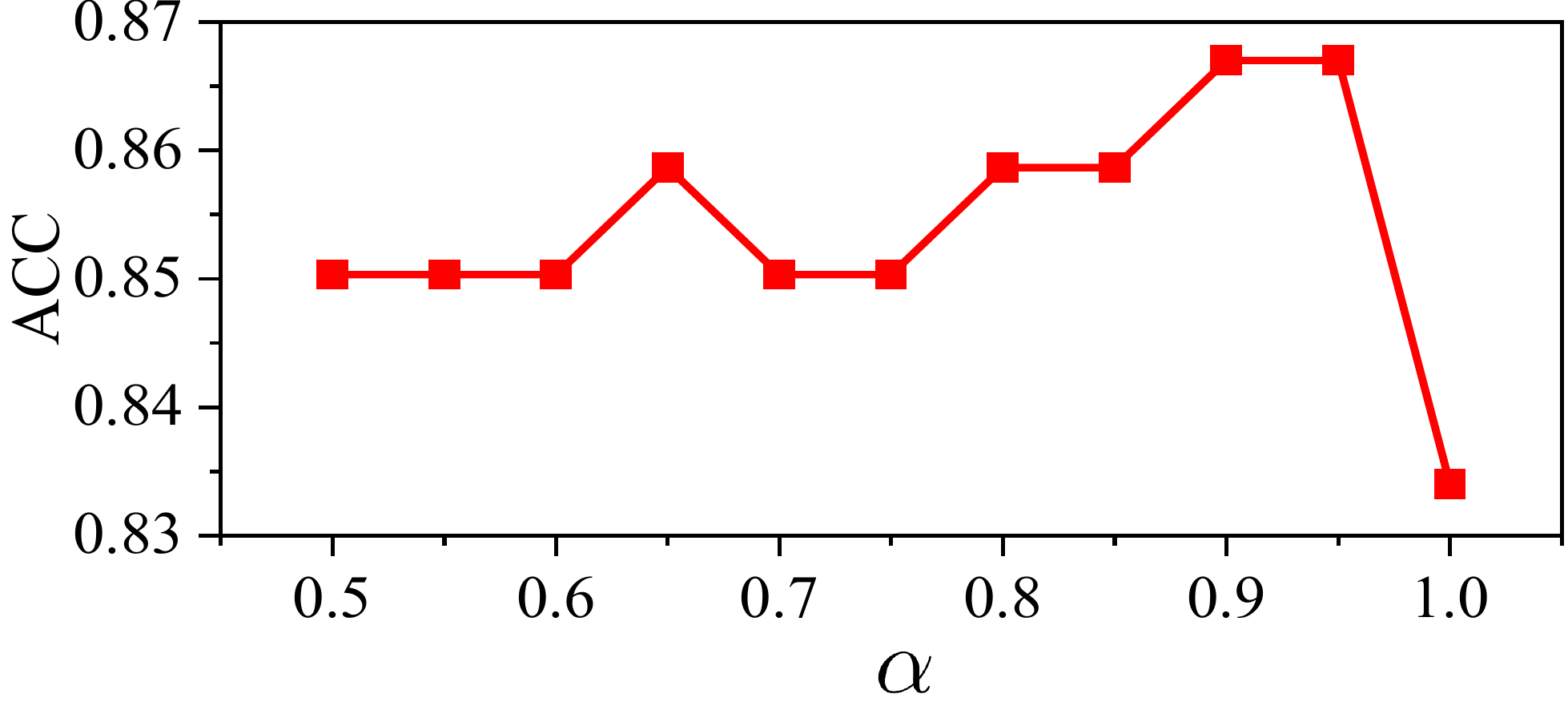}
        \caption{ACC with different \(\alpha\)}
        \label{fig:sensitive4}
    \end{subfigure}
    \caption{Sensitive analysis results on RLT dataset under visual modality. (a) Accuracy with different \(\xi\), (b) Accuracy with different \(\nu\), (c) Accuracy with different \(\eta\), and (d) Accuracy with different \(\alpha\).}
    \label{fig:sensitive}
\end{figure}

\begin{table*}[tb]
    \centering        
     \renewcommand\arraystretch{1.1}
     \scalebox{0.8}{
    \begin{tabular}{m{5.2em}<{\centering}|m{17em}<{\centering}||m{5.2em}<{\centering}|m{17em}<{\centering}}
    \hline
         \textbf{Notations} & \textbf{Description} & \textbf{Notations} & \textbf{Description} \\
         \hline
          \(\mathbb{R}\) & Real number space   &  \(\mathbf{X}^{t}\)/\(\mathbf{X}^{s}\)  &   Feature embeddings of target/source samples   \\
         \hline
          \(F\) &  The number of video frames  &  \(\mathbf{X}^{s,k}\),\(\mathbf{X}^{s,k}_{j}\)  & The feature embeddings of souce samples in \(k\)-th class and the \(j\)-th embedding in \(k\)-th class    \\
         \hline
          \(H\) &  The height of video frames  &  \(J_k\)  &  The number of samples in \(k\)-th category    \\
          \hline
          \(W\) & The width of video frames &  \(d\)  &   The dimension of feature embeddings   \\
         \hline
          \(\mathbf{D}^{t}\)/\(\mathbf{D}^{s}\) &  The target/source dataset  &  \(\mathbf{X}^{t}_{(f)}\)/\newline\(\mathbf{X}^{t}_{(v)}\)/\(\mathbf{X}^{t}_{(a)}\)   &   Feature embeddings of fused ($f$) / visual ($v$) / audio ($a$) modality for target domain samples  \\
         \hline
          \(\mathbf{V}^{t}_i\)/\(\mathbf{V}^{s}_j\) & The \(i\)-th/\(j\)-th video sample in the target/source dataset   & \(\mathcal{F}_1\)/\(\mathcal{F}_2\)  & MLP for feature space mapping  \\
         \hline
          \(N^{t}\)/\(N^{s}\) & The number of samples in the target/source dataset   &  \({\mathbf{X}^{t}}'\)   &    Target feature embeddings after mapping by \(\mathcal{F}_1\)  \\
          \hline
          \(\boldsymbol{y}^{t}\)/\(\boldsymbol{y}^{s}\) & The ground truth labels of samples in target/source dataset  &     $n$     &  The number of given target samples in mini-batch     \\
          \hline
           \(y^{t}_i\)/\(y^{s}_j\)      &   The ground truth label of the \(i\)-th/\(j\)-th sample in target/source dataset    &  $\mathcal{P}$     & The discrete uniform distribution over \(n\) target samples  \\
          \hline
           \(L^{t}\)/\(L^{s}\)     &   The number of target/source dataset categories    &   $\mathcal{Q}$, \(\mathcal{Q}^k\)    &     The discrete uniform distribution over \(L^{s}\) classes and the sample distribution in the \(k\)-th class      \\
          \hline
           \(\mathcal{G}\)     &   Pre-trained model on source dataset    &  \(\delta_{\mathcal{Q}^k}\)/\(\delta_{{\mathbf{X}^{t}_i}'}\)     &    The Dirac function at location \(\mathcal{Q}^k\)/\({\mathbf{X}^{t}_i}'\)       \\
          \hline
           \(\mathcal{E}\)     &   Encoder layer in AFFAKT    &    \(\mathbf{M}\)   &  The cost matrix of high-level OT         \\
          \hline
             \(\mathcal{H}(\cdot)\),\;\(\epsilon\)   &    The entropic regularization function and its factor  &   \(\mathbf{T}\)    &    The transport plan of high-level OT       \\
          \hline
           \(\mathbf{X}^{trans}\)     &   The transferred feature embeddings from source domain    &    \(\Pi(\mathcal{P},\mathcal{Q})\)   &     The joint probability distribution of $\mathcal{P}$ and $\mathcal{Q}$      \\
          \hline
           \(\mathbf{X}^{fused}\)     &  Fused feature embeddings      &   \(\mathbf{M}^{low,k}\)    &   The cost matrix  of the \(k\)-th source class in low-level OT      \\
          \hline
            \(p^k_j\)    &    The importance of the \(j\)-th sample in the \(k\)-th source class     &    $\mathbf{T}^{low,k}$   &    The transport plan of the \(k\)-th source class in low-level OT       \\
          \hline
            \(N_e\)/\(e\)    &   Total/Current training epoch number      &    \(\xi\)/\(\xi'\)   &   The maximum/current weight of \(\mathbf{X}^{trans}\)       \\
          \hline
            $\hat{\boldsymbol{y}}$    &     The prediction of AFFAKT    &   \(\mathcal{F}_3\)    &    Classifier of AFFAKT      \\
          \hline
            \(\mathcal{L}_{ot}\), \(\eta\)    &    The Sinkhorn divergence loss and its factor   &  \(\mathcal{L}_{ce}\)     &     The cross-entropy loss      \\
          \hline
          \(\alpha\)      &  The factor of momentum updating in SRKB module       &    \(\mathbf{B}\)   &    The correlation prototype in SRKB module       \\
          \hline
           \(\mathbb{I}\)     &   The indicator function      &    \(I_i\)   &    The \(\mathbf{B}\) row index for the \(i\)-th target sample       \\
          \hline
           \(\text{std}(\cdot)\)     &   The standard deviation function      &     \(\nu\)    &    The threshold in SRKB module     \\
          \hline
            \(\mathbb{E}[\cdot]\)    &  The expectation function       &   \(\hat{\mathbf{T}}\)    &   The improved transport plan by SRKB module        \\
        \hline
    \end{tabular}
    }
    \caption{Notations and their corresponding descriptions in our method.}
    \label{tab:symbols}
\end{table*}

\cref{fig:sensitive} shows the variation tendencies of AFFAKT's prediction with the changing of parameters, four parameters are tested: (a) \(\xi\), the maximum weight of \(\mathbf{X}^{trans}\), (b) \(\nu\) in Eq. (13), the threshold in the SRKB module, (c) \(\eta\) in Eq. (11), the regularization parameter of Sinkhorn divergence loss, and (d) \(\alpha\) in Eq. (12), the parameter of momentum updating in the SRKB module. The experiments are conducted on the RLT dataset\cite{perez2015deception} with visual modality, the performance is evaluated with ACC metric. When one parameter is tested, other parameters are set as \(\xi=0.2\), \(\nu=0.05\), \(\eta=0.05\), \(\alpha=0.95\), the testing interval of \(\xi\), \(\nu\), \(\eta\), \(\alpha\) are \(0.0\sim 1.0\), \(0.0\sim 0.5\), \(0.0001\sim 1.0\), and \(0.5\sim1.0\), respectively. The experimental settings are same with that in our main text. 

As we can see from \cref{fig:sensitive}, the ACC varies from different values of each hyper-parameter. 
In \cref{fig:sensitive} (a), it is obvious that the optimal ACC is obtained when \(\xi\) is $0.2$, and ACC decreases with the increase of \(\xi\). In our model, 
\(\xi\) denotes the weight of transferred knowledge \(\mathbf{X}^{trans}\) when integrating transferred features and deceptive features. Such results indicate that facial expression as auxiliary information can provide useful knowledge for deception detection, best precision can be achieved when 
a balance between facial expression features and deceptive features is guaranteed.
In \cref{fig:sensitive} (b), the ACC increases till a peak (when \(\nu=0.05\)) and then drops as \(\nu\) increases. 
Especially when \(\nu\geq0.3\), the ACC doesn't change with \(\nu\), which indicates that a suitable threshold \(\nu\) should not be too large to avoid setting all \(\sigma_i\) to $0$. The standard deviation of \(\mathbf{T}_{i}\)  for each sample in deception dataset is expected to be large, since deception should has higher similarity with negative facial expression categories and lower similarity with positive facial expression categories. If the standard deviation of \(\mathbf{T}_{i}\) is small, it means that the calculated \(\mathbf{T}_{i}\) is not useful in transferring proper information from source domain to target domain, the correlation prototype $\mathbf{B}_{I_i}$ would be used as final transport plan \(\hat{\mathbf{T}}_{i}\) instead according to Eq. (13). Therefore, when \(\nu\) is too small, \(\mathbf{T}_{i}\) with small standard deviation could not be discovered effectively, leading to the decrease in ACC.

In \cref{fig:sensitive} (c), the ACC doesn't show a significant changes when \(\eta\leq0.01\). When \(\eta>0.01\), the ACC shows a steep descent as \(\eta\) increases. When \(\eta>0.01\), the supervisory information of distinguishing deception samples from the truthful ones is suppressed, so the discrimination capability of AFFAKT deteriorates.
In \cref{fig:sensitive} (d), the ACC obtains the best performance when the momentum factor \(\alpha=0.9\) and \(\alpha=0.95\), so we take \(\alpha=0.95\) in our experiment. Note that when \(\alpha=1\), the correlation prototype \(\mathbf{B}\) would not be updated.

\section{Interpretability Studies}
\label{sec:interpretability}

In order to further analyze the learned correlation prototype \(\mathbf{B}\), we show the value of \(\mathbf{B}\) in \Cref{tab:prototype} when best accuracy is achieved on the two deception datasets by transferring knowledge from facial expression dataset under three modalities, respectively. According to the results shown in comparison experiments in the main text, the best ACCs are achieved for RLT dataset when the source domains are selected as DFEW, MAFW and MAFW separately under visual, audio and fused modalities. For DOLOS, the best ACCs are achieved when the source domains are selected as DFEW, MAFW and DFEW. Therefore, the corresponding results for each deception dataset with three modalities are shown in \Cref{tab:prototype}. 

There are seven classes (\textit{happy}, \textit{sad}, \textit{neutral}, \textit{angry}, \textit{surprise}, \textit{disgust} and \textit{fear}) in DFEW dataset \cite{jiang2020dfew} and FERV39K dataset \cite{wang2022ferv39k}, eleven classes (\textit{anger}, \textit{disgust}, \textit{fear}, \textit{happiness}, \textit{neutral}, \textit{sad}, \textit{surprise}, \textit{contempt}, \textit{anxiety}, \textit{helpless} and \textit{disappoint}) in MAFW dataset \cite{liu2022mafw}, and two classes (\textit{truthful} and \textit{deceptive}) in both deception datasets. Bold represents the largest three absolute difference between being truthful and deceptive across all categories of source domain in \cref{tab:prototype}. 

As demonstrated by \Cref{tab:prototype} in visual modality, \textit{sad} is significant related to \textit{deceptive} for both RLT and DOLOS, while \textit{happy} is remarkably correlated to \textit{truthful}. Such results are also supported by psychological theory that liars will be less positive and pleasant than truth tellers \cite{depaulo2003cues, zloteanu2020reconsidering, hauch2015computers}. For \textit{neutral} and \textit{fear}, \textit{neutral} has a higher similarity with \textit{truthful} than \textit{deceptive} on both datasets, especially on DOLOS dataset. Besides, \textit{fear} is more correlated with \textit{deceptive} than \textit{truthful} on both datasets. \cite{mathur2020introducing} showed that deceivers in high-stakes situations are likely to associate with \textit{fear}, which is consistent with our results. 

In audio and fused modalities, since that feature representations are extracted from source domain by using visual modality only, the discrepancy between different modalities hinders the establishment of the relationship between facial expressions and deception behavior, resulting in different affective expression correlations between \textit{truthful} and \textit{deceptive} across different modalities. In general, the negative affects are more related with \textit{deception}, such as \textit{fear}, \textit{anxiety}. This phenomenon coincides with previous researches \cite{mathur2020introducing, depaulo2003cues}, which demonstrates that deceivers in high-stakes situations are likely to associate with \textit{fear}. \textit{Neutral} has a generally higher similarity with \textit{truthful} than \textit{deceptive} on both datasets, especially on DOLOS dataset, which is consistent with the results in \cite{zuckerman2014telling}. 

The interpretability studies also confirm that AFFAKT can automatically establish proper relationships between categories of facial expressions and deception datasets, which helps in leveraging transferred knowledge from facial affective dataset during testing phase.

\section{Notations in Our Method}
\label{sec:notations}
\cref{tab:symbols} lists all notations that appear in our method and their corresponding descriptions.

\section{Acknowledgment}
This work is supported by the National Natural Science Foundation of China (No. 62306118 and No. 62207002), Basic and Applied Basic Research Foundation of Guangzhou (2023A04J1682), and the Guangdong Provincial Key Laboratory of Human Digital Twin (2022B1212010004), the Fundamental Research Funds for the Central Universities (2023ZYGXZR105).

\bibliography{main}

\end{document}